\documentclass{article}
\usepackage[preprint]{colm2026_conference}

\usepackage{graphicx}
\usepackage{subcaption}
\usepackage{microtype}
\usepackage{hyperref}
\usepackage{url}
\usepackage{array}
\usepackage{booktabs}
\usepackage{makecell}
\usepackage{tabularx}
\usepackage{multirow}
\usepackage{colortbl}
\usepackage{amsmath}
\usepackage{amssymb}
\usepackage{enumitem}
\usepackage{algorithm}
\usepackage{algorithmic}
\usepackage{tcolorbox}
\tcbuselibrary{breakable, skins}
\usepackage{csquotes}

\newtcolorbox{promptbox}[2][teal!70!black]{%
  title={#2},
  colback=gray!4,
  colframe=#1,
  coltitle=white,
  fonttitle=\bfseries\small,
  fontupper=\ttfamily\scriptsize,
  before upper={\obeylines\setlength{\parskip}{0.5em}\setlength{\parindent}{0pt}},
  breakable,
  enhanced,
  boxrule=0.8pt,
  arc=3pt,
  left=6pt, right=6pt, top=4pt, bottom=4pt,
}
\newcommand{\pvar}[1]{{\color{orange!70!black}\texttt{\{#1\}}}}
\newcommand{\ind}{\hspace*{1.5em}}

\definecolor{darkblue}{rgb}{0, 0, 0.5}
\definecolor{oursrowcolor}{RGB}{229, 239, 255}
\hypersetup{colorlinks=true, citecolor=darkblue, linkcolor=darkblue, urlcolor=darkblue}

\title{SeekerGym: A Benchmark for Reliable Information Seeking}

\author{%
\normalfont%
\hfil
\begin{tabular}{@{}c@{\hspace{4em}}c@{}}
\textbf{Remy Kim}$^*$ & \textbf{Minseung Lee}$^*$ \\
University of Pennsylvania, Curation Labs & Curation Labs \\
\texttt{jkkim123@seas.upenn.edu} & \texttt{mslee@curationlabs.ai} \\[1.5em]
\textbf{Shuo Li} & \textbf{Osbert Bastani}$^\dagger$ \\
Google DeepMind & University of Pennsylvania \\
\texttt{sureli@google.com} & \texttt{obastani@seas.upenn.edu}
\end{tabular}%
\hfil
}

\begin{document}

\maketitle
\begingroup
\renewcommand{\thefootnote}{}
\footnotetext{$^*$\,Equal contribution. \hspace{1em} $^\dagger$\,Corresponding author.}
\endgroup

\begin{abstract}
Despite their substantial successes, AI agents continue to face fundamental challenges in terms of trustworthiness. Consider deep research agents, tasked with searching for information relevant to a given topic---while AI agents can perform effective information retrieval, there is little guarantee regarding the completeness of this information. Gaps in retrieved information can leave biases that mislead users even if the information they are given is correct and relevant. We introduce SeekerGym, a benchmark designed to evaluate the completeness of information retrieved by AI agents. In addition, SeekerGym also measures how well agents quantify their uncertainty in the completeness of their information; if an agent fails to retrieve all relevant information, it is useful for it to at least quantify how much might be missing. At a high level, each task in SeekerGym is a document (e.g., a Wikipedia article), and the AI agent must issue queries to retrieve passages from that document. Intuitively, the document comprehensively covers a topic, so the ability to retrieve its sections directly measures completeness of information retrieval. In addition to Wikipedia, we also consider machine learning survey papers, where the goal is to retrieve relevant sections of a survey paper. We benchmark several models and algorithms; the best approaches retrieve 42.5\% of passages on Wikipedia and 29.2\% on ML Surveys, leaving substantial room for improvement.
\end{abstract}

\section{Introduction}

Users increasingly delegate complex information seeking to agentic systems---deep research agents find relevant sources, software engineering agents retrieve code context, and survey-writing agents discover related literature. Yet, they currently have no way to verify whether the information gathered is complete. Failures are silent---if a deep research agent fails to surface conflicting experimental studies on a topic, it may incorrectly conclude that scientific consensus has been reached, and the user has no basis to question that judgment.

As a consequence, there is a critical need to measure information gathering beyond the accuracy of the information retrieved or the clarity of the way information is presented. We consider two questions. First, \textit{information gathering completeness} measures whether the agent generate queries that retrieve \emph{all} relevant information, not just relevant and accurate information. In practice, collecting all relevant information may not always be feasible; in these cases, it may suffice for users to know that the information is incomplete. Thus, \textit{completeness uncertainty quantification} measures whether the agent accurately estimate the completeness of the information it has gathered.

 
Existing benchmarks for information-seeking agents evaluate end-to-end task performance, entangling the information gathering process with response generation, making it impossible to assess failures of each component in isolation. These benchmarks also neglect uncertainty around completeness, never measuring whether the agent can estimate how much relevant information remains undiscovered. Existing evaluation of uncertainty quantification focuses on verifying a single answer for correctness rather than assessing whether the agent found \textit{all} relevant information. Table~\ref{tab:benchmark-comparison} summarizes these gaps.
 
We introduce \textbf{SeekerGym}, a benchmark environment that shifts evaluation from end-to-end task metrics to direct, capability-level measurement with ground-truth verification. The key challenge is that the full information set is unknown for typical information-gathering tasks. To bridge this gap, we consider documents that comprehensively cover a topic---e.g., encyclopedia articles or scientific surveys. Intuitively, if an agent is gathering information about the topic of such a document, it should be able to retrieve all sections in that document for the information to be considered complete. Thus, each task in SeekerGym is the topic of one such document, and the ground-truth information set is comprised of the sections of that document. Given the topic, an agent must iteratively issue queries to try and retrieve all sections in that document; completeness is measured based on the fraction of sections successfully retrieved, and uncertainty quantification is measured based on the agent's assessment of its own completeness. We construct SeekerGym on two such corpora of documents: Wikipedia articles for broad, general-domain knowledge, and ML Surveys for more specialized scientific knowledge. In summary, our contributions are:
\begin{enumerate}[leftmargin=1.5em, itemsep=0.1em]
\item \textbf{Formalization and benchmark construction.}
We formalize information seeking as two independently measurable capabilities grounded in a POMDP framework, and construct SeekerGym with ground-truth goal sets from two curated corpora and a reproducible evaluation pipeline (Figure~\ref{fig:overview}; Section~\ref{sec:benchmark}).
\item \textbf{Information gathering completeness evaluation.}
We evaluate six models across three belief state representations, finding that document-intrinsic properties dominate performance variance, that belief representation matters at least as much as model identity, and that full exploration history counter-intuitively degrades completeness.
\item \textbf{Completeness uncertainty quantification evaluation.}
We apply conformal prediction to transform raw model completeness estimates into calibrated prediction intervals with finite-sample coverage guarantees, achieving the target 90\% coverage level across all models on both corpora. Models exhibit distinct pre-calibration bias profiles, with larger reasoning models requiring the least correction.
\end{enumerate}

\begin{table}[t]
\caption{Comparison of benchmarks involving information seeking. Existing benchmarks either target specific facts with ground-truth verification or evaluate holistic output via LLM-as-a-judge. None simultaneously isolates information-seeking capability and evaluates termination decisions.}
\label{tab:benchmark-comparison}
\centering
\small
\renewcommand{\arraystretch}{1.3}
\begin{tabular}{lcccc}
\toprule
\textbf{Benchmark} &
\textbf{Scope} &
\textbf{Verifier} &
\textbf{\makecell{Isolated\\Seeking}} &
\textbf{\makecell{Termination\\Eval}} \\
\midrule
HotpotQA, MultihopQA, SimpleQA & \makecell{Specific\\(short-horizon)} & Ground Truth & \texttimes & \texttimes \\
BrowseComp, HLE, xbench & Specific & Ground Truth & \texttimes & \texttimes \\
WebArena, GAIA, SWE-Bench & Task-level & Ground Truth & \texttimes & \texttimes \\
\makecell[l]{DeepResearchGym,\\DeepResearch Bench} & Holistic & LLM-as-Judge & \texttimes & \texttimes \\
\midrule
\rowcolor{oursrowcolor}
\textbf{SeekerGym (Ours)} & \textbf{Holistic} & \textbf{Ground Truth} & \checkmark & \checkmark \\
\bottomrule
\end{tabular}
\end{table}

\textbf{Related work.}
LLM-based agents interleave reasoning with retrieval actions to gather information~\citep{yao2023react, gao2024rag}; this agentic paradigm is increasingly widely used in practice~\citep{openai2025deepresearch}. Recent benchmarks evaluate multi-turn information seeking: WebArena~\citep{zhou2024webarena} and GAIA~\citep{mialon2023gaia} test web navigation and tool use; BrowseComp~\citep{wei2025browsecomp} and FRAMES~\citep{krishna2024frames} require multi-step retrieval over web sources. These benchmarks evaluate end-to-end task success, which entangles query generation with downstream response generation, so failures cannot be attributed to a specific capability. A related line of work addresses \textit{when} to retrieve during generation: Self-RAG~\citep{asai2024selfrag} uses learned reflection tokens to trigger on-demand retrieval, and FLARE~\citep{jiang2023flare} queries when token-level confidence drops. These methods make binary per-step retrieval decisions within a single task and do not address the problem of estimating the \textit{completeness} of gathered information.

Another limitation is the absence of ground truth for information completeness. Existing benchmarks measure whether agents find a single correct answer, not whether they find all relevant information. Concurrent to our work, DeepSearchQA~\citep{gupta2026deepsearchqa} evaluates search comprehensiveness on the open web via outcome-based precision and recall over verifiable answer sets. In contrast, SeekerGym directly measures both information gathering completeness and completeness uncertainty quantification in a controlled document-grounded setting.

In addition, LLM uncertainty
estimation has received growing attention~\citep{geng2024survey}, with
methods ranging from log-probability probes~\citep{kadavath2022language}
and semantic entropy~\citep{kuhn2023semantic} to verbalized
confidence~\citep{tian2023just, lin2022teaching}. While log-probability methods require
logit access and sampling-based methods incur additional computation, verbalized approaches apply to any black-box API and require no architectural modification, making them a practical option for
evaluating a diverse range of models.

\begin{figure}[t]
\centering
\includegraphics[width=0.85\linewidth]{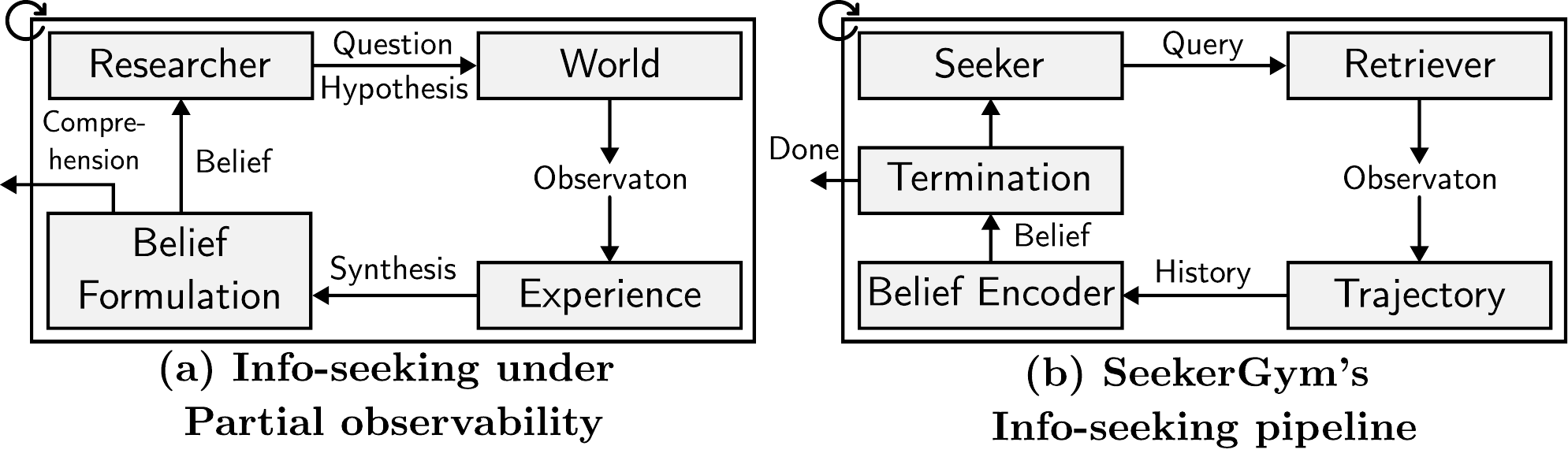}
\caption{\textbf{Information seeking under partial observability and
its SeekerGym instantiation.} (a) Under partial observability, a
researcher iteratively asks questions or forms hypotheses, gathers
evidence, updates a working belief shaped by curiosity and remaining
uncertainty, and decides whether their understanding is sufficient to
stop or whether further exploration is needed. (b) SeekerGym
instantiates this generic loop as a
benchmark pipeline (Algorithm~\ref{alg:evaluation}): the Seeker issues
a query $a_t$, the Retriever returns passages $o_t$, the Belief
Encoder updates the belief state $b_{t+1}$, and the agent either
continues or stops.}
\label{fig:infoseeking_loop}
\end{figure}

\section{SeekerGym: A Benchmark for Information Seeking}
\label{sec:benchmark}

\begin{figure}[t]
\centering
\includegraphics[width=0.9\linewidth]{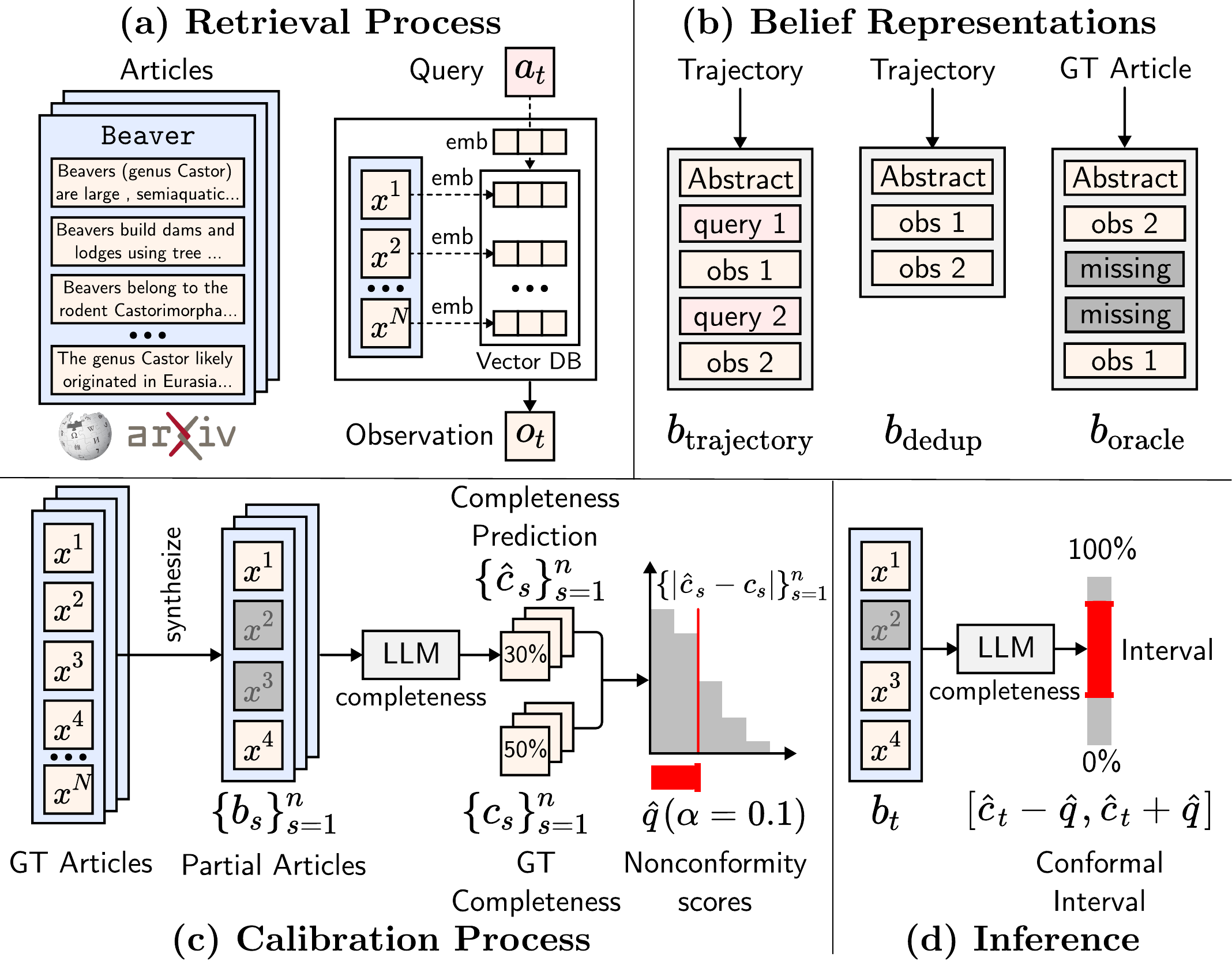}
\caption{\textbf{SeekerGym overview.}
(a)~\textit{Retrieval Process.} Wikipedia articles are parsed into
passages and indexed via $\text{embed}(\cdot)$. Given a query $a_t$,
the Retriever performs vector search over the passage index and
returns all goal passages whose embedding similarity to $a_t$ exceeds
threshold $\theta$ as the observation $o_t$.
(b)~\textit{Belief Representations.} Three representations of the
belief state $b_t$: \textit{raw trajectory} ($b_{\text{trajectory}}$)
preserves the verbatim query--observation history;
\textit{deduplicated trajectory} ($b_{\text{deduplicated}}$) retains
only unique retrieved passages; \textit{oracle belief}
($b_{\text{oracle}}$) uses the ground-truth article structure (GT
Article), placing retrieved passages at their correct positions and
marking unretrieved passages as \texttt{[MISSING]}.
(c)~\textit{Calibration Process} (Algorithms~\ref{alg:snapshots},~\ref{alg:calibration}).
Synthetic belief states $\{b_s\}_{s=1}^{n}$ are fed to a completeness
estimator to obtain predictions $\{\hat{c}_s\}_{s=1}^{n}$;
nonconformity scores $e_s = |c_s - \hat{c}_s|$ against true
completeness ratios $\{c_s\}_{s=1}^{n}$ yield the calibration adjustment
$\hat{q}$. Passages shown in gray boxes are hidden from the estimator and
are used only to define the true completeness ratio.
(d)~\textit{Inference.} Given a deduplicated trajectory belief $b_t$,
the estimator produces $\hat{c}$ from the visible passages only; the
agent terminates (done) when $\hat{c} - \hat{q} \geq \delta$,
guaranteeing $1{-}\alpha$ coverage.}
\label{fig:overview}
\end{figure}

Agentic information seeking is a sequential task: a researcher asks questions or
forms hypotheses, gathers evidence, updates an internal picture of
what remains uncertain, and decides whether they understand the topic
well enough to stop or should continue exploring. Figure~\ref{fig:infoseeking_loop}
illustrates this generic loop and its SeekerGym instantiation.
SeekerGym provides a controlled environment for this process. Given a
target document's abstract, the agent generates natural-language
queries to discover the document's goal passages through
embedding-based retrieval. The benchmark is comprised of a
curated dataset Section~\ref{sec:dataset}) and a POMDP formalization of the seeking
pipeline (Section~\ref{sec:pomdp}). Figure~\ref{fig:overview}
summarizes the SeekerGym pipeline, from retrieval and belief
construction to calibration and inference. SeekerGym evaluates models via two metrics: \textit{information gathering completeness} (Section~\ref{sec:information_gathering}) quantifies the completeness of the information retrieved by the agent within a fixed query budget, and \textit{completeness uncertainty quantification} (Section~\ref{sec:termination_prediction}) quantifies whether the agent can accurately estimate its own completeness.


\subsection{Datasets}
\label{sec:dataset}

SeekerGym accepts any corpus of documents that comprehensively cover
their respective topics, with each document's sections serving as the
goal passages an agent must discover. We instantiate the benchmark
with two complementary corpora: (1) a Wikipedia corpus that
provides broad, community-curated encyclopedic coverage, and (2) an
ML Surveys corpus that tests whether findings transfer to
technical, non-encyclopedic scientific writing. The Wikipedia corpus
contains 120 articles across 15 topic clusters, while the ML Surveys
corpus contains 80 papers across 8 topic areas.
Table~\ref{tab:article_stats} summarizes corpus-level passage
statistics and topical coverage across both corpora.

\begin{figure}[t]
\centering
\includegraphics[width=1.0\linewidth]{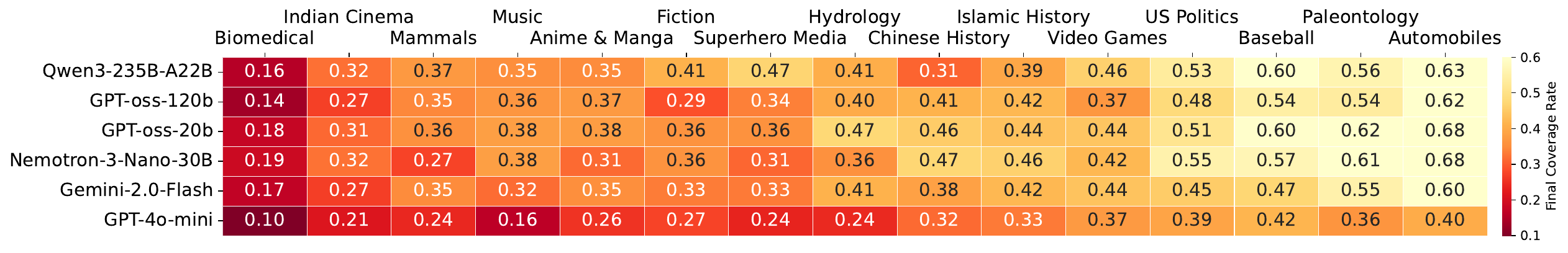}
\caption{Mean final completeness ratio (fraction of goal passages retrieved) per Wikipedia topic cluster and model, using the deduplicated trajectory belief representation (unique retrieved passages as context; see \S\ref{sec:belief_representations}). Each cell averages over 8 Wikipedia articles. Difficulty varies substantially across topics. Dashed line: overall mean across all clusters.}
\label{fig:topic_difficulty}
\end{figure}

\begin{table*}[t]
\centering
\caption{Corpus-level dataset statistics and topical coverage across the two SeekerGym corpora. ML Surveys contains larger goal sets per document on average, while the listed topic areas summarize corpus-level topical coverage.}
\label{tab:article_stats}
\small
\begin{tabularx}{\textwidth}{lXX}
\toprule
 & \textbf{Wikipedia} & \textbf{ML Surveys} \\
\midrule
\makecell[l]{Num passages} & 46.2 & 102.9 \\
\makecell[l]{Tokens / passage} & 164 & 103.0 \\
\midrule
Topic areas &
\textit{Science \& Technology}: Automobiles, Biomedical, Hydrology, Mammals, Paleontology;\newline
\textit{History \& Society}: Baseball, Chinese History, Islamic History, US Politics;\newline
\textit{Arts \& Entertainment}: Anime \& Manga, Fiction, Indian Cinema, Music, Superhero Media, Video Games. &
Graph Learning; Recommender Systems; Diffusion Models; Interpretability / XAI; Federated Learning; Efficient LLMs; LLM Agents; Healthcare \& Medical Imaging. \\
\bottomrule
\end{tabularx}
\end{table*}

Both corpora are constructed through the same
four-stage process:
(1) Parsing converts source documents into structured articles and passages.
(2) Quality filtering removes documents that are too short, fragmented, or otherwise unsuitable for retrieval-based evaluation.
(3) Topic curation identifies coherent topical groups and removes outliers
while preserving benchmark-level diversity.
(3) Final corpus construction
ranks candidates with citation-aware scoring and a short-passage
penalty, then selects the paper-facing corpus.
The exact thresholds,
review procedures, and source-specific cleanup steps differ by domain,
but the shared pipeline ensures that both corpora are built under
comparable quality and diversity criteria (Tables~\ref{tab:pipeline_stats}, \ref{tab:ml_survey_pipeline_stats}, \&~\ref{tab:ml_survey_cluster_selection} in Appendix~\ref{app:dataset_stats}).
We retain 8 Wikipedia articles per cluster and 10 ML survey papers per
topic area, yielding 120 and 80 documents respectively.

\subsection{Information Seeking Pipeline}
\label{sec:pomdp}

We frame information seeking as a partially observable Markov decision process (POMDP)~\citep{kaelbling1998pomdp}. The agent must discover goal passages that it cannot observe directly, using only query responses to infer what remains. Each episode corresponds to a single benchmark document. The agent receives the document's abstract as initial context and iteratively issues queries to uncover goal passages.

\textbf{State.} A state encodes the goal passages $\mathcal{X}^{\text{goal}} = \{x_1, \ldots, x_N\}$ and a binary vector $s^{\text{retrieved}}_t \in \{0,1\}^N$ tracking which passages have been observed. The goal set is fixed within an episode; only $s^{\text{retrieved}}_t$ evolves.

\textbf{Actions.} At each step, the agent issues one or more natural language queries targeting information it believes remains undiscovered.

\textbf{Observations.} Given a query $a$, the environment returns all goal passages whose embedding similarity exceeds threshold $\theta$:
\begin{align*}
O(s, a) = \{x \in \mathcal{X}^{\text{goal}} : \cos(\text{embed}(a), \text{embed}(x)) > \theta\}
\end{align*}
We set $\theta = 0.65$, yielding an empty observation when no passage is sufficiently similar; details on embedding model selection and threshold calibration are provided in Appendix~\ref{app:retrieval}.

\textbf{Transition.} The transition is deterministic: the
retrieved-passage indicator updates as
$s^{\text{retrieved}}_{t+1} = s^{\text{retrieved}}_t \lor
\mathbf{1}[O(s_t, a_t)]$, where $\mathbf{1}[\cdot]$ maps each
returned passage to its index. Retrieved passages accumulate
monotonically: once found, a passage remains observed.

\textbf{Reward.} We define a terminal reward equal to the
completeness ratio at episode end:
$R = |\{i : s_i^{\text{retrieved}} = 1\}| / N$. No intermediate
rewards are used; the agent receives feedback only through
observations. A discount factor $\gamma \in (0, 1]$ can optionally reward early termination; we use $\gamma = 1$ in all experiments (Section~\ref{sec:evaluation_metric}).

\textbf{Belief.} In a standard POMDP the belief is a probability
distribution over hidden states. Here, because the state space is
combinatorial (which subset of $N$ passages has been retrieved) and
observations are natural-language passages, maintaining an explicit
distribution is intractable. We therefore approximate the belief as
a structured textual representation of the exploration history, which
serves as the agent's context for query generation. We denote this
approximate belief $b_t$; it accumulates retrieved passages and
(depending on the representation) query history over the episode.
Different belief representations compress or restructure the trajectory
$\tau_t = (a_1, o_1, \ldots, a_t, o_t)$ in different ways
(see Section~\ref{sec:belief_representations}).

\textbf{Termination.} The agent does not explicitly decide when to stop. Instead, we ask the agent to quantify their uncertainty in the completeness of their information (detailed in Section~\ref{sec:termination_prediction}); the user can decide when to terminate execution based on this completeness estimate.

\textbf{Horizon.} Each episode runs for at most $M$ steps
(finite horizon) or terminates early when the calibrated stopping
criterion is met (Section~\ref{sec:termination_prediction}).



\subsection{Information Gathering Completeness}
\label{sec:information_gathering}
\label{sec:evaluation_metric}

The core measure of seeking effectiveness is whether the agent retrieves
the passages it needs. Under a fixed budget of $M$~steps with
$K$~queries each, the completeness ratio captures the fraction retrieved---analogous
to recall in information retrieval:
\begin{align*}
c = \frac{|\{i : s_i^{\text{retrieved}} = 1\}|}{|\mathcal{X}^{\text{goal}}|}
\end{align*}
All agents operate under the same query budget, so this ratio directly reflects information-seeking capability. We compare completeness ratios across belief representations and model families.
Table~\ref{tab:query_effectiveness} previews the main completeness results across both corpora; we analyze these in Section~\ref{sec:experiments}. The completeness ratio can be extended with a discount factor $\gamma \in (0, 1]$ to reward early termination: $\text{Score} = c \times \gamma^{t^* - M}$, where $t^*$ is the termination step and $M$ is the maximum step budget. When $\gamma = 1$ (our setting), this reduces to the plain completeness ratio. Values $\gamma < 1$ assign a bonus of $\gamma^{t^* - M} > 1$ to episodes that terminate before the budget is exhausted, incentivizing efficiency alongside completeness.

\begin{table}[t]
\caption{Information-gathering completeness by model and belief representation. Left: Wikipedia (3 seeds). Right: ML Surveys (3 seeds). Across all models on both corpora, completeness follows the same ordering: raw trajectory (Traj.) $<$ deduplicated trajectory (Dedup.) $<$ oracle belief (Oracle), while absolute completeness is lower on ML Surveys. The oracle belief representation uses privileged gap information and serves as an upper bound. $\dagger$: reasoning/thinking model.}
\label{tab:query_effectiveness}
\centering
\small
\begin{tabular}{lcccccc}
\toprule
 & \multicolumn{3}{c}{Wikipedia} & \multicolumn{3}{c}{ML Surveys} \\
\cmidrule(lr){2-4}\cmidrule(lr){5-7}
Model & Traj. & Dedup. & Oracle & Traj. & Dedup. & Oracle \\
\midrule
Qwen3-235B-A22B$^\dagger$ & \cellcolor{blue!10}27.5\% & \cellcolor{blue!21}41.0\% & \cellcolor{blue!35}57.6\% & \cellcolor{blue!5}11.6\% & \cellcolor{blue!18}23.1\% & \cellcolor{blue!33}37.2\% \\
GPT-oss-120b$^\dagger$ & \cellcolor{blue!7}24.2\% & \cellcolor{blue!19}38.5\% & \cellcolor{blue!31}52.8\% & \cellcolor{blue!7}13.4\% & \cellcolor{blue!17}22.2\% & \cellcolor{blue!34}37.9\% \\
GPT-oss-20b$^\dagger$ & \cellcolor{blue!12}30.2\% & \cellcolor{blue!22}42.5\% & \cellcolor{blue!29}50.8\% & \cellcolor{blue!7}13.9\% & \cellcolor{blue!19}24.1\% & \cellcolor{blue!35}38.2\% \\
Nemotron-3-Nano-30B$^\dagger$ & \cellcolor{blue!7}23.9\% & \cellcolor{blue!21}40.5\% & \cellcolor{blue!23}43.7\% & \cellcolor{blue!8}15.1\% & \cellcolor{blue!24}28.9\% & \cellcolor{blue!29}33.8\% \\
Gemini-2.0-Flash & \cellcolor{blue!9}26.1\% & \cellcolor{blue!19}37.9\% & \cellcolor{blue!22}42.2\% & \cellcolor{blue!13}19.1\% & \cellcolor{blue!24}29.2\% & \cellcolor{blue!25}30.1\% \\
GPT-4o-mini & \cellcolor{blue!5}20.6\% & \cellcolor{blue!11}28.0\% & \cellcolor{blue!17}35.6\% & \cellcolor{blue!10}16.7\% & \cellcolor{blue!19}24.5\% & \cellcolor{blue!24}28.8\% \\
\bottomrule
\end{tabular}
\end{table}

\subsection{Completeness Uncertainty Quantification}
\label{sec:termination_prediction}

A reliable agent must provide information to help the user decide when to stop execution. We evaluate this by prompting
the agent to estimate its current completeness ratio---i.e., the fraction of
goal passages retrieved so far. This metric can help the user make well-informed termination decisions. However, LLMs exhibit systematic biases in self-assessment: a model may
report 50\% completeness when the true rate is 70\%. Calibration
corrects this using held-out data where the true completeness ratio is known.
We apply conformal prediction~\citep{angelopoulos2021gentle}, which
yields prediction intervals with finite-sample guarantees. We define
the true completeness ratio as
$c = |\textsc{Retrieved}| / N \in [0, 1]$ and evaluate on synthetic belief states constructed offline
(Algorithm~\ref{alg:snapshots},
Appendix~\ref{app:synthetic_beliefs}). Because all models receive
identical belief states, performance differences reflect purely how well
each model estimates completeness, not how well it generates queries.

In more detail, an estimator produces a predicted completeness ratio
$\hat{c}$ for each synthetic belief state. We calibrate these
estimates via conformal prediction~\citep{angelopoulos2021gentle}: on
the calibration set, we compute nonconformity scores
$e_i = |c_i - \hat{c}_i|$ and set the calibration adjustment $\hat{q}$
to the $\lceil(n+1)(1-\alpha)\rceil / n$ quantile, yielding prediction
intervals $[\hat{c} - \hat{q},\; \hat{c} + \hat{q}]$ with coverage
guarantee
$P(c \in [\hat{c} - \hat{q},\; \hat{c} + \hat{q}]) \geq 1 - \alpha$.
Background on conformal
prediction is in Appendix~\ref{app:conformal_background};
Algorithm~\ref{alg:calibration}
in Appendix~\ref{app:calibration_algorithm} summarizes the full
procedure.
We evaluate calibration quality through the interval half-width
$\hat{q}$; smaller $\hat{q}$ indicates tighter calibration since the base estimator requires less correction.

\section{Experiments}
\label{sec:experiments}

\begin{figure}[t]
\centering
\includegraphics[width=\linewidth]{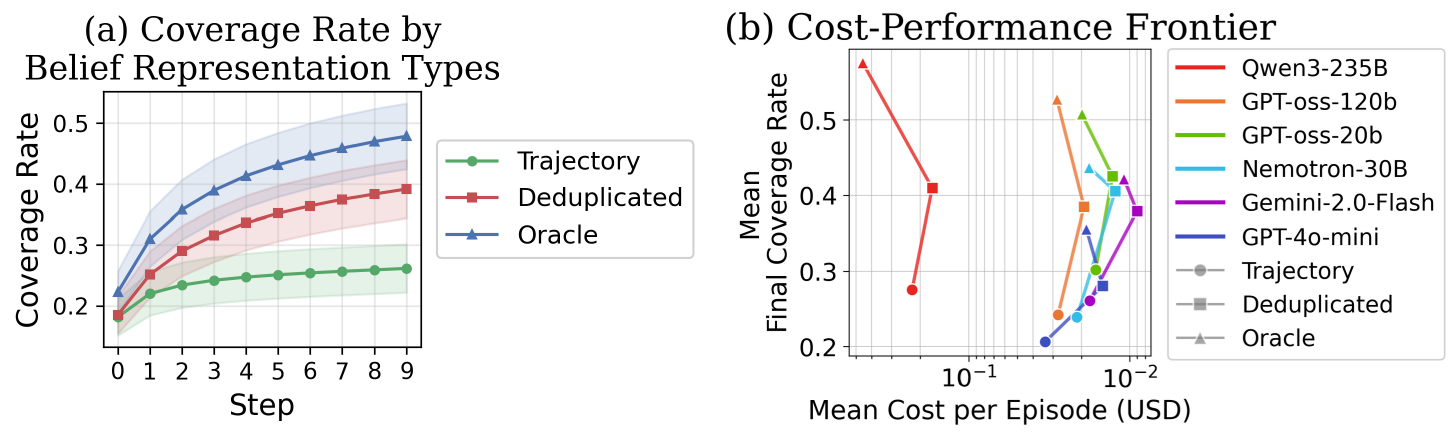}
\caption{(a) Completeness ratio by belief representation types on Wikipedia,
averaged across all 6 models and 3 seeds per article ($n{=}120$
Wikipedia articles). Shaded bands show $\pm 0.3$ std across articles.
Among the three belief representations, oracle belief provides
privileged gap information and serves as an upper bound; raw
trajectory plateaus early while deduplicated trajectory and oracle
belief continue to improve. Per-model breakdowns appear in
Appendix~\ref{fig:recall_steps_all}. (b) Cost-performance frontier on
Wikipedia: cost per episode vs.\ mean completeness ratio across all
(model, belief representation) configurations. Because the x-axis is
flipped, configurations nearest the upper right are the cheapest and
highest-completeness. Figure~\ref{fig:ml_survey_recall_and_cost} in
Appendix~\ref{app:information_gathering} shows the corresponding ML
Surveys frontier.}
\label{fig:recall_and_cost}
\end{figure}

\subsection{Setup}
\label{sec:belief_representations}

\textbf{Models.} We evaluate six models spanning reasoning and standard instruction-following families. Open-source reasoning models include \textbf{GPT-oss-120b}, \textbf{GPT-oss-20b}, \textbf{Qwen3-235B-A22B-Thinking}, and \textbf{Nemotron-3-Nano-30B-A3B}. Proprietary models include \textbf{GPT-4o-mini} and \textbf{Gemini-2.0-Flash}. This selection spans model scale and generation strategy, enabling us to examine whether chain-of-thought reasoning and model size independently improve seeking behavior. We select smaller proprietary models (GPT-4o-mini, Gemini-2.0-Flash) to balance evaluation cost against benchmark scale (each (model, belief) configuration requires 120 Wikipedia episodes and 80 ML Surveys episodes of multi-turn interaction), while the open-source models span a wider parameter range to test scaling effects. Appendix~\ref{app:hyperparameters} provides hyperparameter details and the full evaluation protocol.

\textbf{Belief representations.} The belief representation $b_t$ determines what context the agent receives when generating queries. We
evaluate three representations: \textit{raw trajectory} preserves the
full query--observation history verbatim; \textit{deduplicated
trajectory} retains only unique retrieved passages, eliminating
duplicates across queries; and \textit{oracle belief} provides the
article's full section structure with retrieved passages in place and
unretrieved passages marked as \texttt{[MISSING]} placeholders, serving
as an upper bound via privileged structural information unavailable at
deployment. SeekerGym supports arbitrary alternative representations.
See Appendix~\ref{app:prompts} for full prompt formats.

\textbf{Completeness estimation.} For completeness uncertainty quantification, the agent is prompted to judge completeness of the current
retrieved passages, returning a predicted completeness ratio
$\hat{c} \in [0, 1]$ as a point estimate of the true ratio
$c = |\textsc{Retrieved}| / N$. This estimate is calibrated via
conformal prediction (Section~\ref{sec:termination_prediction}).
SeekerGym supports alternative estimation methods. See Appendix~\ref{app:prompts} for the full prompt format.

\begin{table}[t]
\caption{Completeness-ratio results across corpora. $\hat{q}$: conformal half-width, where smaller values indicate less calibration correction. $R^2$: raw out-of-sample coefficient of determination. Only GPT-oss-120b and GPT-oss-20b achieve positive raw $R^2$ on both corpora, indicating that conformal calibration remains necessary for most models.}
\label{tab:completeness_ratio}
\centering
\small
\begin{tabular}{lcccc}
\toprule
 & \multicolumn{2}{c}{Wikipedia} & \multicolumn{2}{c}{ML Surveys} \\
\cmidrule(lr){2-3}\cmidrule(lr){4-5}
Model & $\hat{q}$ & $R^2$ & $\hat{q}$ & $R^2$ \\
\midrule
Qwen3-235B-A22B & \cellcolor{blue!21}$0.560 \pm 0.002$ & $-0.529 \pm 0.042$ & \cellcolor{blue!20}$0.557 \pm 0.006$ & $-0.434 \pm 0.014$ \\
GPT-oss-120b & \cellcolor{blue!35}$0.378 \pm 0.007$ & $0.426 \pm 0.016$ & \cellcolor{blue!35}$0.367 \pm 0.011$ & $0.453 \pm 0.029$ \\
GPT-oss-20b & \cellcolor{blue!29}$0.462 \pm 0.004$ & $0.077 \pm 0.023$ & \cellcolor{blue!26}$0.477 \pm 0.008$ & $0.044 \pm 0.024$ \\
Nemotron-3-Nano-30B & \cellcolor{blue!20}$0.573 \pm 0.006$ & $-0.378 \pm 0.048$ & \cellcolor{blue!20}$0.560 \pm 0.006$ & $-0.350 \pm 0.012$ \\
GPT-4o-mini & \cellcolor{blue!25}$0.514 \pm 0.003$ & $-0.131 \pm 0.014$ & \cellcolor{blue!23}$0.525 \pm 0.009$ & $-0.214 \pm 0.022$ \\
Gemini-2.0-Flash & \cellcolor{blue!7}$0.742 \pm 0.001$ & $-1.553 \pm 0.023$ & \cellcolor{blue!7}$0.724 \pm 0.005$ & $-1.408 \pm 0.012$ \\
\bottomrule
\end{tabular}
\end{table}

\subsection{Information Gathering Completeness}

Table~\ref{tab:query_effectiveness} reports completeness ratios across models and belief representations. A mixed-effects ANOVA (Table~\ref{tab:anova_variance_decomposition}, Appendix~\ref{app:information_gathering}) confirms that document-intrinsic properties account for the majority of completeness-ratio variance, while belief representation and model identity have comparable but smaller effects; belief manipulation produces a wider effect range than model selection. The same pattern holds on ML Surveys, as shown in the ML Surveys columns of Table~\ref{tab:anova_variance_decomposition}. We investigate four questions.

\textbf{RQ1: Why does oracle belief outperform deduplicated trajectory?} The oracle belief representation consistently achieves the highest completeness ratio, as expected from its privileged access to positional gap information. We hypothesize that explicit gap information enables more targeted query generation, directing the model toward specific missing content rather than requiring it to infer gaps from context. We provide chain-of-thought length analysis in Appendix~\ref{app:oracle_analysis}.

\textbf{RQ2: Why does raw trajectory underperform deduplicated trajectory?} Despite preserving richer exploration history, raw trajectory consistently underperforms deduplicated trajectory. Figure~\ref{fig:raw_traj_analysis}(a) (Appendix~\ref{app:information_gathering}) shows that raw trajectory accumulates substantially more context over time, while deduplicated trajectory remains much more compact as later steps increasingly retrieve already-seen passages. This growing context burden likely introduces redundant context that exceeds models' effective attention spans~\citep{liu2024lost}. Figure~\ref{fig:raw_traj_analysis}(b) (Appendix~\ref{app:information_gathering}) shows the behavioral consequence: raw trajectory's query diversity collapses after only a few steps, while deduplicated trajectory maintains higher diversity and more sustained retrieval progress.

\textbf{RQ3: How does model scale and reasoning affect coverage?} Larger reasoning models (Qwen3-235B, GPT-oss-120b) generally outperform smaller models (Nemotron-3-Nano-30B) and non-reasoning models (GPT-4o-mini, Gemini-2.0-Flash). One counter-intuitive finding: GPT-oss-20b outperforms GPT-oss-120b on both raw trajectory and deduplicated trajectory belief representations. Table~\ref{tab:reasoning_tokens} in Appendix~\ref{app:oracle_analysis} reveals that GPT-oss-20b produces substantially longer reasoning chains than GPT-oss-120b across all belief representations, suggesting that reasoning depth, not parameter count, is the primary driver of query quality. 


\textbf{RQ4: How does topic affect difficulty?} Figure~\ref{fig:topic_difficulty} reveals substantial variation in completeness ratio across Wikipedia topic clusters, with some topics consistently difficult across models. One exception is Qwen3-235B-A22B's low performance on Chinese History, which is driven by Alibaba Cloud's content moderation filter (Appendix~\ref{app:content_filter}). Figure~\ref{fig:ml_survey_topic_difficulty} (Appendix~\ref{app:information_gathering}) shows that ML Surveys topic areas exhibit a similar difficulty pattern.

\subsection{Completeness Uncertainty Quantification}


All completeness uncertainty quantification experiments use the deduplicated trajectory belief representation with the completeness estimation method described above. We apply conformal prediction to the completeness estimates at target coverage $1 - \alpha = 90\%$. To account for sensitivity to data split randomness, we repeat the evaluation over 3 random splits, average within each split over 10 random seeds, and report mean $\pm$ std across the resulting split-level means~\citep{shao1993linear}. Table~\ref{tab:completeness_ratio} presents the main comparison results for completeness uncertainty quantification across the two corpora, including model-wise $\hat{q}$ and raw $R^2$.

\textbf{RQ5: Can LLMs estimate exploration completeness?}
On Wikipedia, completeness estimation produces calibrated intervals with
$\hat{q} = 0.38$--$0.74$ across models
(Table~\ref{tab:completeness_ratio}). On ML Surveys, the corresponding
range is $\hat{q} = 0.37$--$0.72$. GPT-oss-120b achieves the tightest
intervals on both corpora ($\hat{q} = 0.38$, $R^2 = 0.43$ on
Wikipedia; $\hat{q} = 0.37$, $R^2 = 0.45$ on ML Surveys), indicating
that its raw completeness scores track the true completeness ratio
reasonably well. However, four of six models yield negative $R^2$
values on both corpora, meaning their raw estimates are less accurate
than predicting the mean completeness ratio. Across both corpora, the raw
estimates also exhibit a shared early-stage optimism bias: when true
completeness ratio is near zero, all models tend to overpredict
the completeness ratio. Conformal calibration absorbs this inaccuracy and still
achieves the target 90\% coverage, but at the cost of wider intervals
for weaker estimators. These results suggest that while the
completeness framing produces estimates amenable to
calibration~\citep{kadavath2022language,
xiong2024llmsexpress}, reliable stopping decisions benefit from external
calibration mechanisms rather than model estimation alone, with larger
reasoning models requiring the least correction.
See Figures~\ref{fig:scatter_direct_grid}
and~\ref{fig:ml_survey_scatter_direct_grid}
(Appendix~\ref{app:termination_prediction}) for scatter plots of
true completeness ratio (x-axis) vs.\ predicted completeness ratio (y-axis) on
both corpora, and Figures~\ref{fig:nonconformity_histogram_grid}
and~\ref{fig:ml_survey_nonconformity_histogram_grid} for the
corresponding nonconformity-score distributions underlying conformal
calibration.

\textbf{RQ6: How do models differ in the bias profile of raw completeness estimation before conformal calibration?}
Figures~\ref{fig:scatter_direct_grid}
and~\ref{fig:ml_survey_scatter_direct_grid}
(Appendix~\ref{app:termination_prediction}) indicate that
pre-calibration errors are structured in model-specific ways before
conformal calibration. Points aligned with the diagonal would indicate
well-calibrated raw estimates; systematic deviations reveal
model-specific bias profiles.
Gemini-2.0-Flash often assigns near-complete scores even when true
completeness ratio is still very low. GPT-4o-mini rarely predicts
near-complete states, even when the underlying article is nearly
exhausted. GPT-oss-20b and GPT-oss-120b stay closest to the diagonal,
so their raw estimates track true completeness ratio more closely than
those of the other models. Qwen3-235B-A22B-Thinking tends to
overestimate completeness at intermediate completeness levels while still
recognizing near-complete states accurately.

\section{Conclusion}

We presented SeekerGym, a benchmark that isolates and directly measures two information-seeking capabilities, information gathering completeness and completeness uncertainty quantification, with ground-truth verification across two corpora: 120 Wikipedia articles spanning 15 topic clusters and 80 ML Surveys papers spanning 8 topic areas. Our experiments surface three implications for the design of information-seeking agents. First, the dominance of \textbf{document-intrinsic properties} over both model and belief representation choice suggests that benchmark evaluations using only a handful of topics may be unreliable; topic-diverse test suites are a prerequisite for meaningful comparison. Second, the finding that \textbf{belief representation} manipulation produces a wider effect range than model selection indicates that advances in how agents structure retrieved context may yield larger practical gains than scaling model parameters alone. Third, \textbf{completeness estimation} paired with conformal calibration produces usable stopping signals: the model's raw assessment of exploration completeness, while imperfect, can be transformed into calibrated prediction intervals with finite-sample guarantees, suggesting that practical termination benefits from pairing LLM judgment with external calibration mechanisms.



\bibliography{references}
\bibliographystyle{colm2026_conference}

\appendix

\section{Embedding and Retrieval Details}
\label{app:retrieval}

\subsection{Embedding Model}

The observation function returns passages whose embedding similarity to the query exceeds a threshold $\theta$. All benchmark evaluations use \textbf{OpenAI text-embedding-3-large} (3072 dimensions). Embedding-based retrieval suits our evaluation goal: articulating specific information gaps requires full natural language sentences, not keyword bags. Embedding models are also self-contained: unlike BM25, they require no corpus-level statistics.

\subsection{Threshold Selection}

The similarity threshold $\theta$ controls task difficulty. Too low, and retrieval becomes trivial, as passages match regardless of query quality. Too high, and the task is unsolvable, since even well-crafted queries fail to retrieve.

\begin{figure}[ht]
\centering
\begin{subfigure}[t]{0.48\linewidth}
    \centering
    \includegraphics[width=\linewidth]{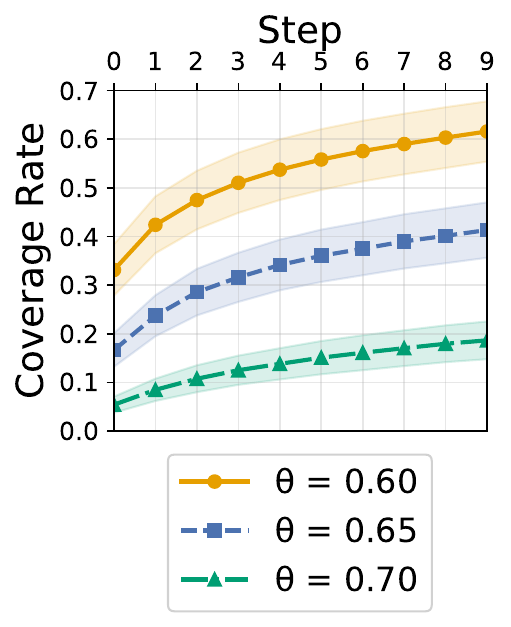}
    \caption{Completeness ratio over exploration steps for Qwen3-235B-A22B (deduplicated trajectory) at three similarity thresholds. Results averaged over 3 seeds across all 120 Wikipedia articles; shaded bands show $\pm 0.3$ std.}
\end{subfigure}
\hfill
\begin{subfigure}[t]{0.48\linewidth}
    \centering
    \input{figures/threshold_examples.tex}
    \caption{Qualitative example for the query \texttt{``Maserati MC12 0-60 mph acceleration time and curb weight''} against the \textit{Maserati MC12} article. Each colored box shows passages retained at a given threshold ($\theta$); bracketed scores are cosine similarities. At $\theta{=}0.60$ (orange) three passages match; at $\theta{=}0.65$ (blue, our default) only the highest-scoring passage survives; at $\theta{=}0.70$ (green) even that passage is filtered out (score $0.697 < 0.70$), illustrating that too-high thresholds starve the agent of signal.}
\end{subfigure}
\caption{Threshold sensitivity analysis. (a) Aggregate completeness ratio across thresholds. (b) A concrete retrieval example from the \textit{Maserati MC12} article illustrating the precision--recall tradeoff at each threshold.}
\label{fig:threshold_ablation}
\end{figure}

We select $\theta$ empirically by evaluating Qwen3-235B-A22B with deduplicated trajectory across candidate thresholds $\theta \in \{0.60, 0.65, 0.70\}$. Figure~\ref{fig:threshold_ablation}(a) shows completeness ratio over exploration steps for each threshold. At $\theta = 0.60$, retrieval is permissive and the agent reaches the highest final completeness, but many retrieved passages are only loosely related to the query. At $\theta = 0.70$, even well-crafted queries frequently return nothing, starving the agent of information. We set $\theta = 0.65$, which balances retrieval quality and quantity: strict enough to avoid noise, permissive enough that the agent can still make progress.

Figure~\ref{fig:threshold_ablation}(b) illustrates this tradeoff with a concrete example. The query \texttt{"Maserati MC12 0-60 mph acceleration time and curb weight"} retrieves three passages at $\theta = 0.60$: one directly about acceleration stats (score 0.697), one about body design and weight distribution (0.628), and one about engine specs (0.600). At $\theta = 0.65$, only the most relevant passage survives. At $\theta = 0.70$, even that passage is filtered out (0.697 $<$ 0.70), leaving the agent with nothing despite a well-targeted query.

\section{Dataset Details}
\label{app:dataset_stats}

\subsection{Detailed Curation Pipeline}

SeekerGym constructs both corpora through multi-stage curation
pipelines tailored to their source documents while preserving a shared
four-stage structure. We describe the Wikipedia and ML Surveys
pipelines below using parallel stage names.

\subsubsection{Wikipedia Curation Pipeline}

SeekerGym curates Wikipedia articles through a four-stage pipeline.

\begin{table}[h]
\centering
\caption{Dataset statistics across Wikipedia pipeline stages. Values are median [Q1--Q3]. $^*$Tokens/passage estimated as $\lfloor\text{len(text)}/4\rfloor$. Filtering and selection remove short, low-quality articles and retain longer, highly cited ones.}
\label{tab:pipeline_stats}
\begin{tabular}{lrrrr}
\toprule
Stage & \# Articles & Citations & Passages / Article & Tokens / Passage$^*$ \\
\midrule
Parsing & 4,755,590 & 5 [2--17] & 7 [3--18] & 30 [9--92] \\
Quality Filtering & 55,504 & 46 [26--105] & 21 [15--32] & 148 [99--211] \\
Topic Curation & 5,276 & 47 [27--109] & 21 [15--32] & 150 [100--216] \\
Final Corpus & 120 & 494 [294--1,055] & 46 [39--54] & 170 [116--236] \\
\bottomrule
\end{tabular}
\end{table}

\paragraph{Parsing.} We parse a full English Wikipedia dump into
structured articles, extracting section hierarchies and splitting body
text into passage-level units. We discard redirects, disambiguation
pages, and articles whose abstracts fall outside 64--1024 tokens. We
also compute inbound citation counts from the full inter-article link
graph.

\paragraph{Quality filtering.} We require minimum 12 passages and a
short passage ratio below 0.1, where a passage is considered short if
it contains fewer than 48 tokens (rejecting articles where more than
10\% of passages are short). These bounds address a practical
constraint: embedding similarity between natural language queries
(typically short, fixed-length) and passages degrades when passages
are too short (insufficient semantic content) or too long (diluted
signal). Filtering for moderate-length passages improves retrieval
reliability. We additionally require at least 16 inbound citations
from other Wikipedia articles and 4 outbound references as a proxy for
quality; highly-linked articles tend to be well-maintained and cover
notable topics.

\paragraph{Topic curation.} We embed article abstracts using
\texttt{text-embedding-3-large} and apply UMAP (output dim 1536)
followed by DBSCAN ($\varepsilon = 0.16$, min\_samples$= 160$,
Euclidean metric) to identify coherent topic groups and remove
outliers that do not fit established clusters. Clustering parameters
were selected via grid search.

\paragraph{Human review within topic curation.} Semantic clustering
produces 35 candidate topic clusters. Researchers review cluster
samples and apply four exclusion criteria:
\begin{enumerate}
    \item \textbf{Template-dominated structure.} Clusters where articles follow near-identical section layouts, testing template recall rather than information seeking (e.g., Tropical Cyclones \& Hurricane Seasons, US Highways \& Interstate Routes, Radio \& Television Stations, Rail \& Transit Systems).
    \item \textbf{Narrow biographical lists.} Clusters consisting primarily of person articles within a single narrow activity, offering limited topical diversity within the cluster (e.g., Cricket Players, Ice Hockey Players, American Football Players, Professional Wrestling, Film Actresses).
    \item \textbf{Narrow geographic or institutional scope.} Clusters tied to a specific nation's niche institutions with poor generalizability (e.g., British Soap Opera Characters, Romanian Literature, NYC Buildings, Commonwealth Politicians, German \& Royal Navy Warships, Military Officers).
    \item \textbf{Redundancy with a selected cluster.} Clusters semantically overlapping with an already-selected cluster (e.g., Japanese JRPGs $\leftrightarrow$ Video Games, Popular Music Releases $\leftrightarrow$ Music Genres).
\end{enumerate}
After curation, we retain 15 topic clusters spanning science \& technology, history \& society, and arts \& entertainment. Table~\ref{tab:cluster_selection} lists all 35 candidate clusters with their inclusion status.

\begin{table*}[t]
\centering
\caption{Wikipedia topic-cluster selection for the final benchmark. Candidate clusters are grouped by final status and sorted alphabetically within each group. Exclusion reasons are (T) template-dominated structure, (B) narrow biographical list, (G) narrow geographic or institutional scope, and (R) redundancy with a selected cluster.}
\label{tab:cluster_selection}
\small
\begin{tabular}{@{}l l@{}}
\toprule
Status & Topic Cluster \\
\midrule
\multirow[t]{15}{*}{\textbf{Included}} & American Politicians \& Political History \\
& Automobiles \& Motor Vehicles \\
& Biomedical Sciences \& Molecular Biology \\
& Chinese History \& Historical Figures \\
& English-language Fiction \& Literature \\
& Hydrology \& Water Infrastructure \\
& Indian Cinema \& Film Industry \\
& Islamic Civilization \& Historical Figures \\
& Japanese Anime \& Manga \\
& Major League Baseball \\
& Mammals \& Mammalian Biology \\
& Music Genres \& Subcultures \\
& Paleontology \& Mesozoic Reptiles \\
& Superhero Media (Marvel \& DC) \\
& Video Games (Western \& Indie) \\
\addlinespace
\multirow[t]{5}{*}{\textbf{Excluded (B)}} & American \& British Film Actresses \\
& American Football Players \& Coaches \\
& Cricket Players \& Figures \\
& Ice Hockey Players \& Figures \\
& Professional Wrestling Events \& Promotions \\
\addlinespace
\multirow[t]{8}{*}{\textbf{Excluded (G)}} & British Soap Opera Characters \\
& Commonwealth \& Irish Politicians \\
& French \& Allied Naval Vessels \\
& German \& Austro-Hungarian Warships \\
& Military Officers \& Commanders \\
& Notable Buildings \& Structures \\
& Romanian Literature, Culture \& Politics \\
& Royal Navy \& Commonwealth Warships \\
\addlinespace
\multirow[t]{3}{*}{\textbf{Excluded (R)}} & Bird Species \& Ornithology \\
& Japanese Video Games \& JRPGs \\
& Popular Music Releases \\
\addlinespace
\multirow[t]{4}{*}{\textbf{Excluded (T)}} & North American Radio \& Television Stations \\
& Rail \& Public Transit Systems \\
& Tropical Cyclones \& Hurricane Seasons \\
& United States Highways \& Interstate Routes \\
\bottomrule
\end{tabular}
\end{table*}

\paragraph{Final corpus construction.} Final article selection
combines popularity with a short-passage-ratio penalty:
\begin{align}
\text{score} = \log(\text{citations} + 1) \times (1 - \text{short passage ratio})
\end{align}
The logarithm dampens citation counts, preventing extremely popular articles from dominating. The short-passage penalty ensures articles with many short or empty passages score poorly even if popular. We select the top 8 articles per cluster, yielding 120 articles total (15 clusters $\times$ 8 articles).

\subsubsection{ML Surveys Curation Pipeline}

SeekerGym curates the ML Surveys corpus through a parallel four-stage
pipeline.

\begin{table}[h]
\centering
\caption{Dataset statistics across ML Surveys pipeline stages. Values are median [Q1--Q3]. $^*$Tokens/passage estimated as $\lfloor\mathrm{len(text)}/4\rfloor$. Filtering and topic curation retain structurally complete survey papers, and final corpus construction selects the most highly cited papers within curated topic areas.}
\label{tab:ml_survey_pipeline_stats}
\begin{tabular}{lrrrr}
\toprule
Stage & \# Articles & Citations & Passages / Article & Tokens / Passage$^*$ \\
\midrule
Parsing & 4,532 & 8 [2--27] & 97 [61--141] & 135 [77--215] \\
Quality Filtering & 3,848 & 9 [2--28] & 97 [66--135] & 144 [84--226] \\
Topic Curation & 336 & 13 [4--39] & 101 [82--123] & 151 [87--236] \\
Final Corpus & 80 & 88 [49--133] & 102 [81--125] & 149 [88--229] \\
\bottomrule
\end{tabular}
\end{table}

\paragraph{Parsing.} We begin from HTML-available survey papers and
parse each article into a structured representation with section titles
and paragraph-level body passages. The parser performs light
extraction-time cleanup, including whitespace normalization, removal
of obvious accessibility residue, and exclusion of low-value fragments
such as caption-dominated blocks and boilerplate-like text. This stage
yields the first paragraph-bearing corpus used by downstream filtering
and selection.

\paragraph{Quality filtering.} We next filter parsed articles using
article-level quality signals derived from the extracted paragraph
corpus. Concretely, we require more than 30 and fewer than 237 body
paragraphs, at least 8 sections, a short passage ratio no greater than
0.4195, and a caption-like paragraph ratio below 0.05. These
constraints remove structurally incomplete or excessively fragmented
articles while preserving broad topical coverage across survey papers.

\paragraph{Topic curation.} We embed article abstracts using
\texttt{text-embedding-3-large} and apply UMAP followed by DBSCAN to
identify dense topic regions in the quality-filtered pool. In the
canonical run, clustering uses a 1536-dimensional UMAP projection with
cosine distance, followed by DBSCAN with $\varepsilon = 0.16$ and
\texttt{min\_samples} $= 64$. We then export cluster dossiers
containing titles and abstracts for external review, assign topic
labels, and remove cluster-level outliers that do not fit the dominant
topic of the cluster. The resulting curated pool defines the topic
areas used in the paper-facing ML Surveys corpus. For the ML Surveys
corpus, the paper-facing presentation uses topic areas rather than
internal cluster IDs. Table~\ref{tab:ml_survey_cluster_selection}
summarizes the 8 included topic areas and 5 excluded topic areas
considered during final topic selection.

\begin{table*}[t]
\centering
\caption{ML Surveys topic selection for the final benchmark. Starting from cluster topics in the broader computer-science survey set, we retain broad ML areas suitable for document-level exploration. Excluded topics are either outside this ML-focused scope or too niche relative to the selected set.}
\label{tab:ml_survey_cluster_selection}
\small
\begin{tabular}{@{}l l@{}}
\toprule
Status & Topic Area \\
\midrule
\multirow[t]{8}{*}{\textbf{Included}} & Diffusion Models \\
& Efficient LLMs \\
& Federated Learning \\
& Graph Learning \\
& Healthcare \& Medical Imaging \\
& Interpretability / XAI \\
& LLM Agents \\
& Recommender Systems \\
\addlinespace
\multirow[t]{5}{*}{\textbf{Excluded}} & Blockchain \\
& LLM Security, Safety \& Privacy \\
& Reasoning in LLMs \\
& Software Engineering \\
& Wireless AI / 6G \\
\bottomrule
\end{tabular}
\end{table*}

\paragraph{Final corpus construction.} From the curated topic pool, we
rank papers within each topic area using citation-aware scoring with a
short-passage penalty:
\begin{align}
\text{score} = \log(\text{citations} + 1) \times (1 - \text{short passage ratio}).
\end{align}
We retain the top 10 papers per topic area, yielding 80 articles
across 8 included topic areas. We then apply LLM-based passage
normalization and a final cleanup pass. These stages are
retrieval-oriented rather than paraphrastic: they remove citation and
math serialization artifacts, metadata residue, HTML boilerplate, and
dangling prose caused by omitted display equations, while preserving
the original meaning and paragraph structure as much as possible.

\subsection{Corpus Statistics}

Table~\ref{tab:article_stats} in the main text summarizes corpus-level
passage statistics and topic coverage across both SeekerGym corpora.
We report corpus-level means rather than cluster-level averages because
overall benchmark scale and topic coverage are the primary dataset
characteristics used in the paper.


\subsection{Synthetic Belief Generation}
\label{app:synthetic_beliefs}

Evaluating completeness uncertainty quantification requires partially-completed belief
states with known true completeness ratios. We construct these for both
corpora through controlled random sampling rather than extracting them
from actual agent trajectories, for three reasons. First, \textit{decoupling}: real
trajectories tie belief state content to the specific agent's query
quality; synthetic states ensure all models are evaluated on identical
inputs, isolating estimation ability from query generation ability.
Second, \textit{completeness-ratio diversity}: agents retrieve easy passages first
and plateau at model-specific ceilings, concentrating belief states in
narrow completeness-ratio bands; random sampling ensures belief states span the
full range from low to high completeness. Third,
\textit{exchangeability}: conformal prediction requires exchangeable
calibration samples; agent trajectories exhibit sequential correlation,
whereas independently sampled belief states satisfy this assumption by
construction.

For each article with $N$ passages, we sample belief states by
partitioning the remaining-passage axis into regular intervals and
drawing one state per interval using the procedure in
Algorithm~\ref{alg:snapshots}. We use $\Delta = 10$ for Wikipedia and
$\Delta = 15$ for ML Surveys, reflecting their different article-length
distributions. For example, an article with 30 passages and
$\Delta = 10$ yields three synthetic belief states spanning low-, mid-,
and high-coverage regimes. The resulting pools are split 50\%/50\% into
calibration and test sets. On Wikipedia, this yields 595 belief states
in total (297 calibration, 298 test; mean 5.0 per article). On ML
Surveys, the same procedure yields 578 belief states in total before
splitting (mean 7.2 per article), which are then divided evenly between
calibration and test.

\begin{algorithm}[H]
\caption{Synthetic Belief Generation}
\label{alg:snapshots}
\begin{algorithmic}[1]
\REQUIRE Article $d$ with $N$ passages $\{x_1, \ldots, x_N\}$, interval size $\Delta$
\STATE $\textsc{Beliefs} \gets []$
\STATE Partition remaining-passage counts $\{0, \ldots, N-1\}$ into contiguous bins of width $\Delta$
\FOR{each remaining-count bin $B$}
    \STATE $n_{\textsc{remaining}} \gets$ sample uniformly from $B$
    \STATE $n_{\textsc{retrieved}} \gets N - n_{\textsc{remaining}}$
    \STATE $\textsc{Retrieved} \gets$ random subset of $\{x_1, \ldots, x_N\}$ with $|\textsc{Retrieved}| = n_{\textsc{retrieved}}$
    \STATE $c \gets n_{\textsc{retrieved}} / N$ \COMMENT{True completeness ratio}
    \STATE $b \gets$ concatenate passages in $\textsc{Retrieved}$ (deduplicated trajectory format)
    \STATE Append $(b, c)$ to $\textsc{Beliefs}$
\ENDFOR
\RETURN $\textsc{Beliefs}$
\end{algorithmic}
\end{algorithm}

\section{Information Seeking Analysis}
\label{app:information_gathering}

\subsection{Variance Decomposition}
\label{app:variance_decomposition}

Table~\ref{tab:anova_variance_decomposition} reports a mixed-effects ANOVA
with article as random intercept across both corpora. Document-intrinsic
properties account for the majority of variance in coverage on both Wikipedia
($\eta^2 = 0.711$) and ML Surveys ($\eta^2 = 0.631$), confirming that topic
selection is the primary driver of benchmark difficulty. Among experimental
factors, belief representation and model identity explain comparatively
smaller variance, while the belief-model interaction is more pronounced on
ML Surveys ($\eta^2 = 0.046$) than on Wikipedia ($\eta^2 = 0.019$). On
Wikipedia, belief manipulation still spans a wider effect range (21.7~pp
from raw trajectory to oracle belief) than model selection (14.0~pp from GPT-4o-mini
to Qwen3-235B-A22B).

\begin{table}[t]
\caption{Type~III ANOVA variance decomposition across corpora. Document-intrinsic properties dominate completeness-ratio variance in both corpora, while belief representation and model identity have materially smaller effects. We report degrees of freedom, F statistics, and effect sizes for direct comparison; all non-residual effects are statistically significant in both corpora.}
\label{tab:anova_variance_decomposition}
\centering
\small
\begin{tabular}{lcccccccc}
\toprule
 & \multicolumn{4}{c}{Wikipedia} & \multicolumn{4}{c}{ML Surveys} \\
\cmidrule(lr){2-5}\cmidrule(lr){6-9}
Factor & \textbf{df} & $\boldsymbol{F}$ & $\boldsymbol{\eta^2}$ & $\boldsymbol{\eta^2_p}$ & \textbf{df} & $\boldsymbol{F}$ & $\boldsymbol{\eta^2}$ & $\boldsymbol{\eta^2_p}$ \\
\midrule
\textbf{Belief Representation} & 2 & 277.0 & \cellcolor{blue!5}0.020 & 0.080 & 2 & 213.7 & \cellcolor{blue!6}0.023 & 0.092 \\
\textbf{Model Identity} & 5 & 138.7 & \cellcolor{blue!6}0.024 & 0.099 & 5 & 38.9 & \cellcolor{blue!5}0.011 & 0.044 \\
Belief $\times$ Model & 10 & 53.7 & \cellcolor{blue!5}0.019 & 0.078 & 10 & 83.1 & \cellcolor{blue!6}0.046 & 0.164 \\
Document & 119 & 169.4 & \cellcolor{blue!35}0.711 & 0.761 & 119 & 145.4 & \cellcolor{blue!32}0.631 & 0.731 \\
Residual & 6343 & --- & \cellcolor{blue!14}0.224 & 0.500 & 4223 & --- & \cellcolor{blue!14}0.232 & 0.500 \\
\bottomrule
\end{tabular}
\end{table}

\subsection{Why does oracle belief enable better queries?}
\label{app:oracle_analysis}

On Wikipedia, the oracle belief representation consistently outperforms deduplicated trajectory in information gathering completeness. We investigate three complementary explanations, then compare the resulting reasoning-length patterns across both corpora.

\begin{figure}[h]
\centering
\includegraphics[width=0.65\linewidth]{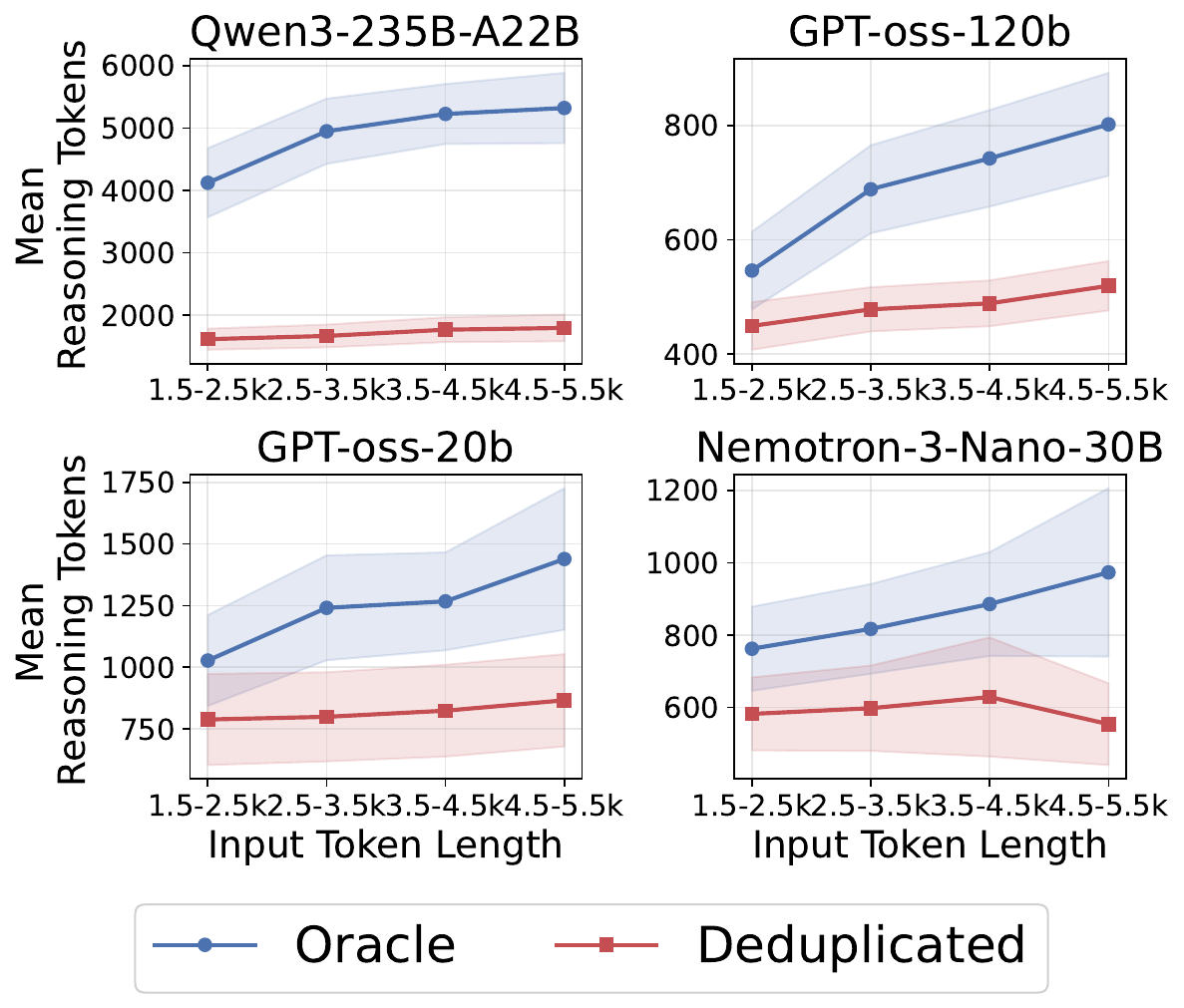}
\caption{Mean reasoning token count per query step on Wikipedia under oracle belief vs.\ deduplicated trajectory (reasoning models only). Compared with deduplicated trajectory, oracle belief elicits longer chains despite minimal additional prompt tokens from positional gap information.}
\label{fig:reasoning_length}
\end{figure}

\paragraph{1. Reasoning chain length (reasoning models only).} We compare the output reasoning token count per query step under oracle belief vs.\ deduplicated trajectory, restricted to reasoning models (GPT-oss-120b, GPT-oss-20b, Qwen3-235B-A22B, Nemotron-3-Nano-30B). The oracle belief representation produces substantially longer reasoning chains despite adding only a negligible number of prompt tokens; the positional gap information (\texttt{<missing id=pN>}) contribute $O(N)$ tokens where $N$ is the number of gaps, a trivially small fraction of the observed reasoning length difference. This rules out prompt inflation as an explanation~(b) and supports the interpretation that oracle belief's structural scaffolding enables deeper, more focused reasoning about each specific gap~(a), consistent with findings that longer reasoning chains correlate with better answer quality~\citep{muennighoff2025s1}. Figure~\ref{fig:reasoning_length} shows this effect across reasoning models.

\begin{table}[t]
\caption{Mean reasoning token length ($\pm$ std) by model and belief representation across corpora. Across both corpora, oracle belief generally elicits longer reasoning chains than deduplicated trajectory, with the largest increases for Qwen3-235B and GPT-oss-20b. Only reasoning models are shown. Abbrev.: Qwen3-235B = Qwen3-235B-A22B; Nemotron-30B = Nemotron-3-Nano-30B.}
\label{tab:reasoning_tokens}
\centering
\small
\begin{tabular}{lcccccc}
\toprule
 & \multicolumn{3}{c}{Wikipedia} & \multicolumn{3}{c}{ML Surveys} \\
\cmidrule(lr){2-4}\cmidrule(lr){5-7}
Model & Traj. & Dedup. & Oracle & Traj. & Dedup. & Oracle \\
\midrule
Qwen3-235B & \cellcolor{blue!16}2180 {\tiny ($\pm$ 1096)} & \cellcolor{blue!13}1675 {\tiny ($\pm$ 669)} & \cellcolor{blue!35}4936 {\tiny ($\pm$ 1898)} & \cellcolor{blue!17}1689 {\tiny ($\pm$ 856)} & \cellcolor{blue!14}1400 {\tiny ($\pm$ 656)} & \cellcolor{blue!35}3540 {\tiny ($\pm$ 2165)} \\
GPT-oss-120b & \cellcolor{blue!5}404 {\tiny ($\pm$ 149)} & \cellcolor{blue!5}453 {\tiny ($\pm$ 144)} & \cellcolor{blue!6}695 {\tiny ($\pm$ 306)} & \cellcolor{blue!5}388 {\tiny ($\pm$ 189)} & \cellcolor{blue!5}375 {\tiny ($\pm$ 147)} & \cellcolor{blue!5}417 {\tiny ($\pm$ 241)} \\
GPT-oss-20b & \cellcolor{blue!5}548 {\tiny ($\pm$ 335)} & \cellcolor{blue!7}758 {\tiny ($\pm$ 586)} & \cellcolor{blue!10}1247 {\tiny ($\pm$ 816)} & \cellcolor{blue!7}667 {\tiny ($\pm$ 326)} & \cellcolor{blue!8}743 {\tiny ($\pm$ 482)} & \cellcolor{blue!10}1004 {\tiny ($\pm$ 552)} \\
Nemotron-30B & \cellcolor{blue!6}698 {\tiny ($\pm$ 730)} & \cellcolor{blue!6}565 {\tiny ($\pm$ 397)} & \cellcolor{blue!8}875 {\tiny ($\pm$ 555)} & \cellcolor{blue!6}558 {\tiny ($\pm$ 243)} & \cellcolor{blue!5}394 {\tiny ($\pm$ 173)} & \cellcolor{blue!7}640 {\tiny ($\pm$ 328)} \\
\bottomrule
\end{tabular}
\end{table}

\begin{figure}[h]
\centering
\includegraphics[width=0.65\linewidth]{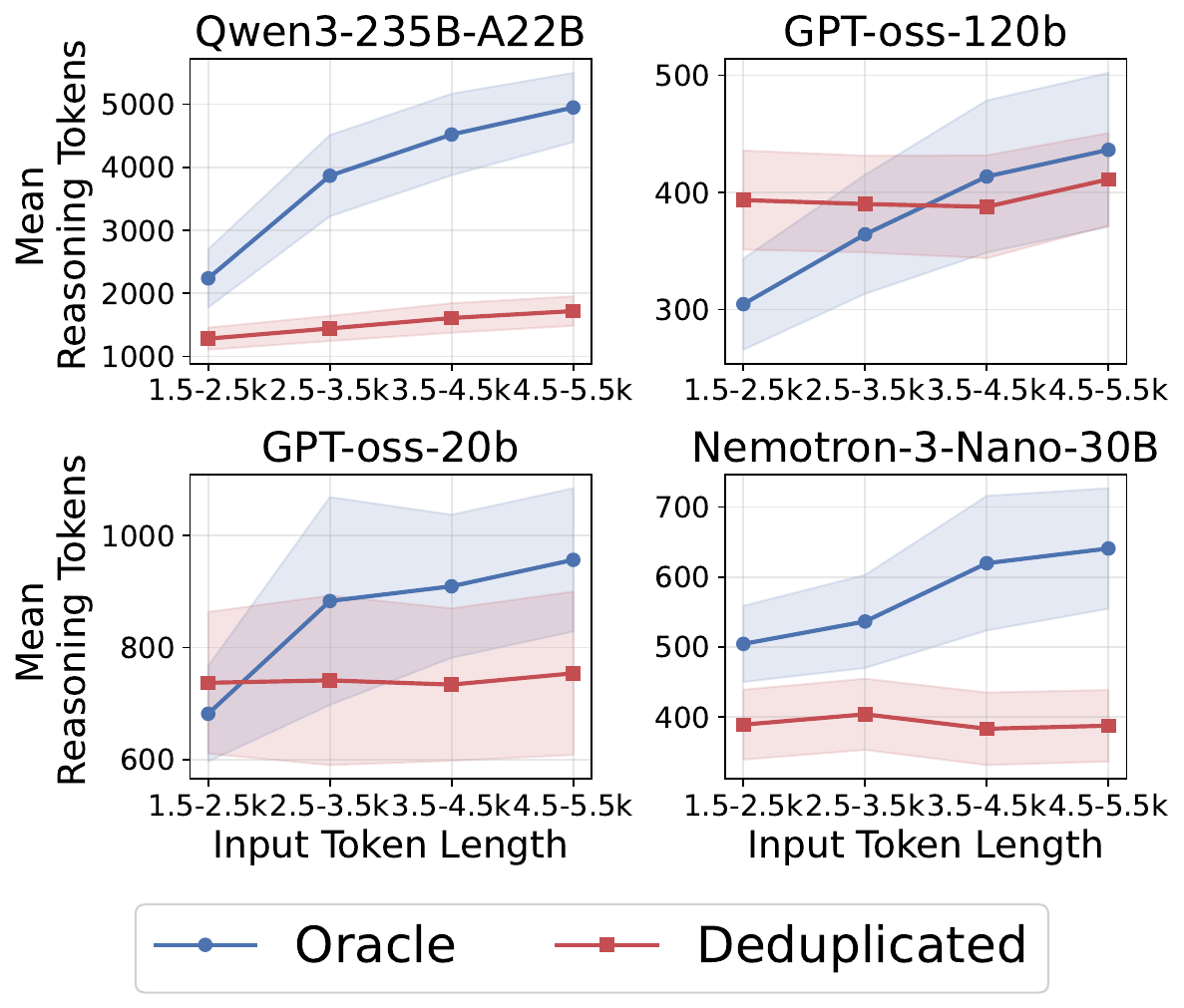}
\caption{Mean reasoning token count per query step on ML Surveys under oracle belief vs.\ deduplicated trajectory (reasoning models only). Compared with deduplicated trajectory, oracle belief again elicits longer chains despite minimal additional prompt tokens from positional gap information.}
\label{fig:ml_survey_reasoning_length}
\end{figure}

\paragraph{2. Subgoal structure as query scope reduction.} The oracle belief representation reveals not just that information is missing but \textit{where}: the location of each gap relative to found passages. This structural information allows the model to scope each query to a specific target region rather than searching globally. Without it, deduplicated trajectory forces the model to simultaneously infer \textit{what} is missing and \textit{where} to look, a harder joint inference problem. This suggests that future belief representations could close the oracle belief--deduplicated trajectory gap by estimating gap locations from retrieved passages, without requiring privileged access.


\subsection{Completeness Ratio over Exploration Steps: Per-Model Breakdown}

\begin{figure}[ht]
\centering
\includegraphics[width=\linewidth]{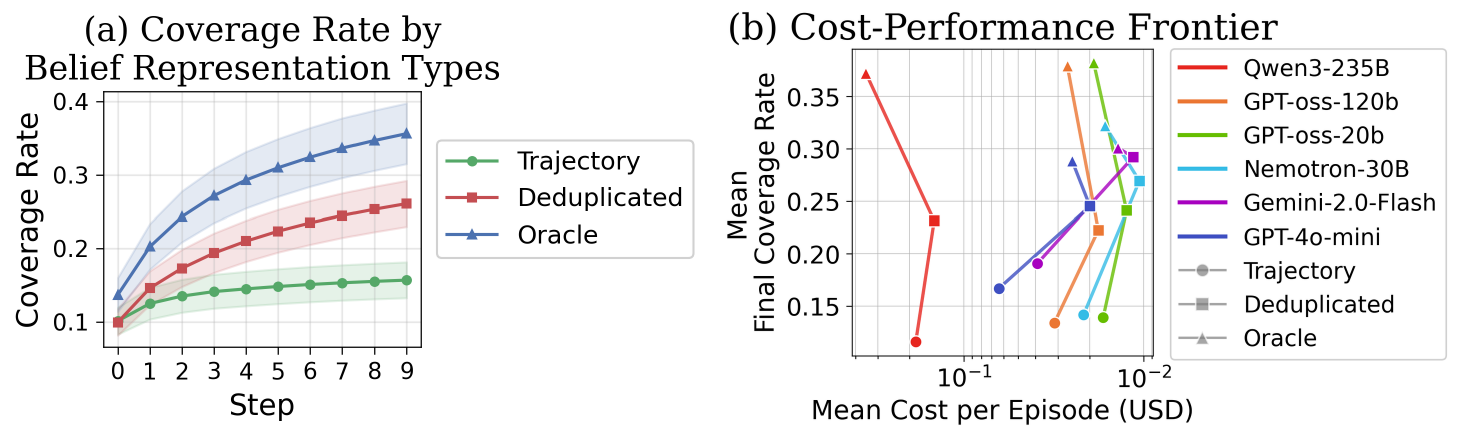}
\caption{ML Surveys counterpart to Figure~\ref{fig:recall_and_cost}.
(a) Completeness ratio by belief representation types on ML Surveys.
Shaded bands show $\pm 0.3$ std across episodes. As on Wikipedia, raw
trajectory plateaus early while deduplicated trajectory and oracle
belief continue to improve. (b) Cost-performance frontier on ML
Surveys. Because the x-axis is flipped, configurations nearest the
upper right are the cheapest and highest-completeness.}
\label{fig:ml_survey_recall_and_cost}
\end{figure}

\begin{figure}[ht]
\centering
\includegraphics[width=\linewidth]{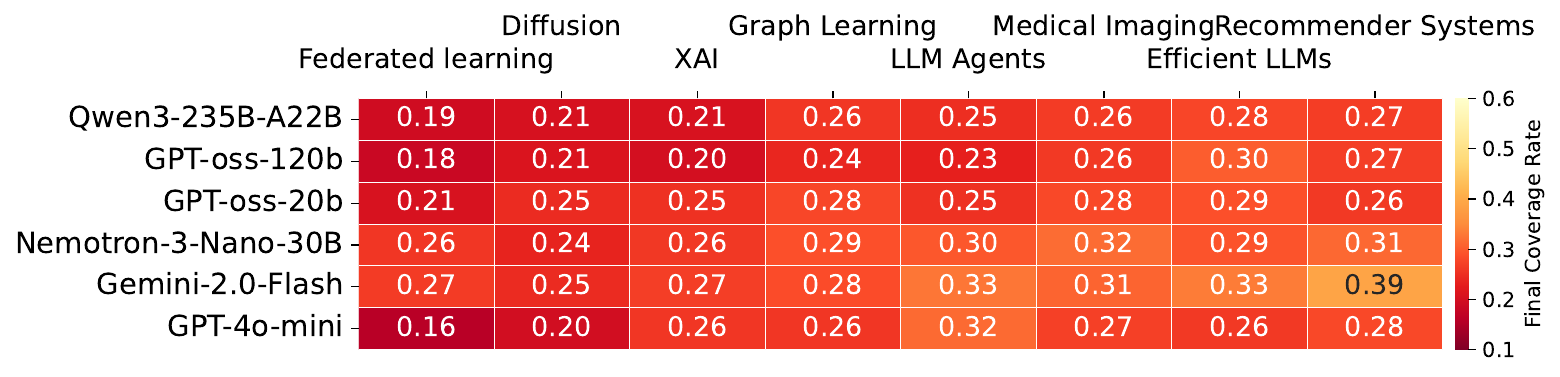}
\caption{Mean final completeness ratio by topic area and model on ML Surveys, using the deduplicated trajectory belief representation (unique retrieved passages as context; see \S\ref{sec:belief_representations}). Difficulty varies substantially across topic areas. Dashed line: overall mean across all topic areas.}
\label{fig:ml_survey_topic_difficulty}
\end{figure}

\begin{figure}[ht]
\centering
\includegraphics[width=\linewidth]{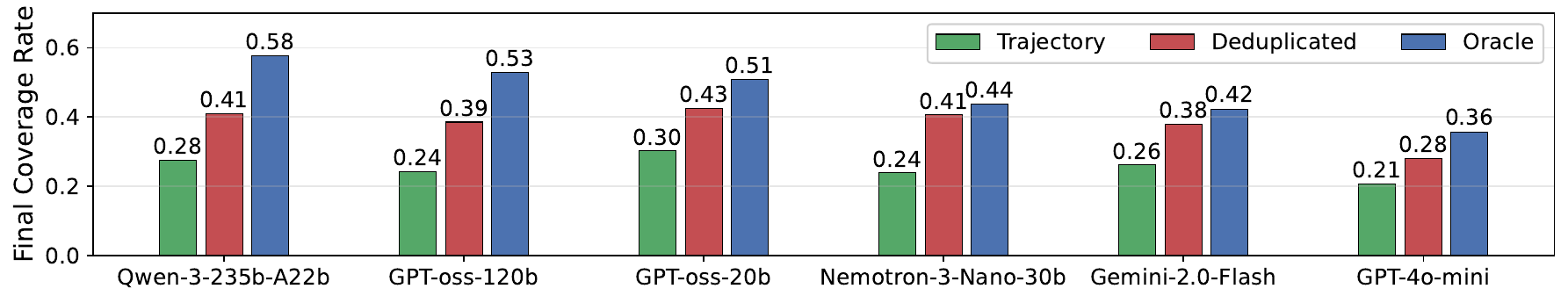}
\caption{Final completeness ratio by model and belief representation on Wikipedia. Models sorted by deduplicated trajectory completeness ratio. The raw trajectory $<$ deduplicated trajectory $<$ oracle belief ordering holds consistently across all models.}
\label{fig:recall_bar_per_model}
\end{figure}

\begin{figure}[ht]
\centering
\includegraphics[width=\linewidth]{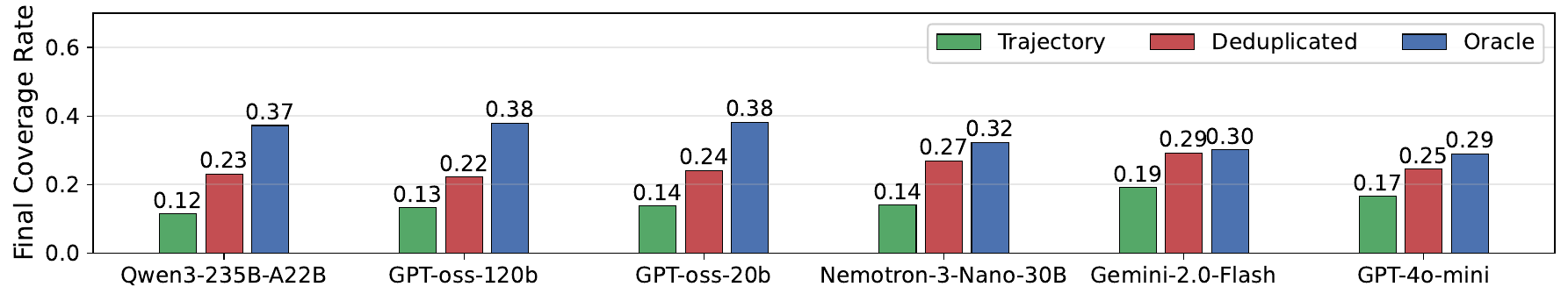}
\caption{Final completeness ratio by model and belief representation on ML Surveys. Models sorted by deduplicated trajectory completeness ratio. The raw trajectory $<$ deduplicated trajectory $<$ oracle belief ordering remains visible across models.}
\label{fig:ml_survey_recall_bar_per_model}
\end{figure}

\begin{figure}[ht]
\centering
\includegraphics[width=0.9\linewidth]{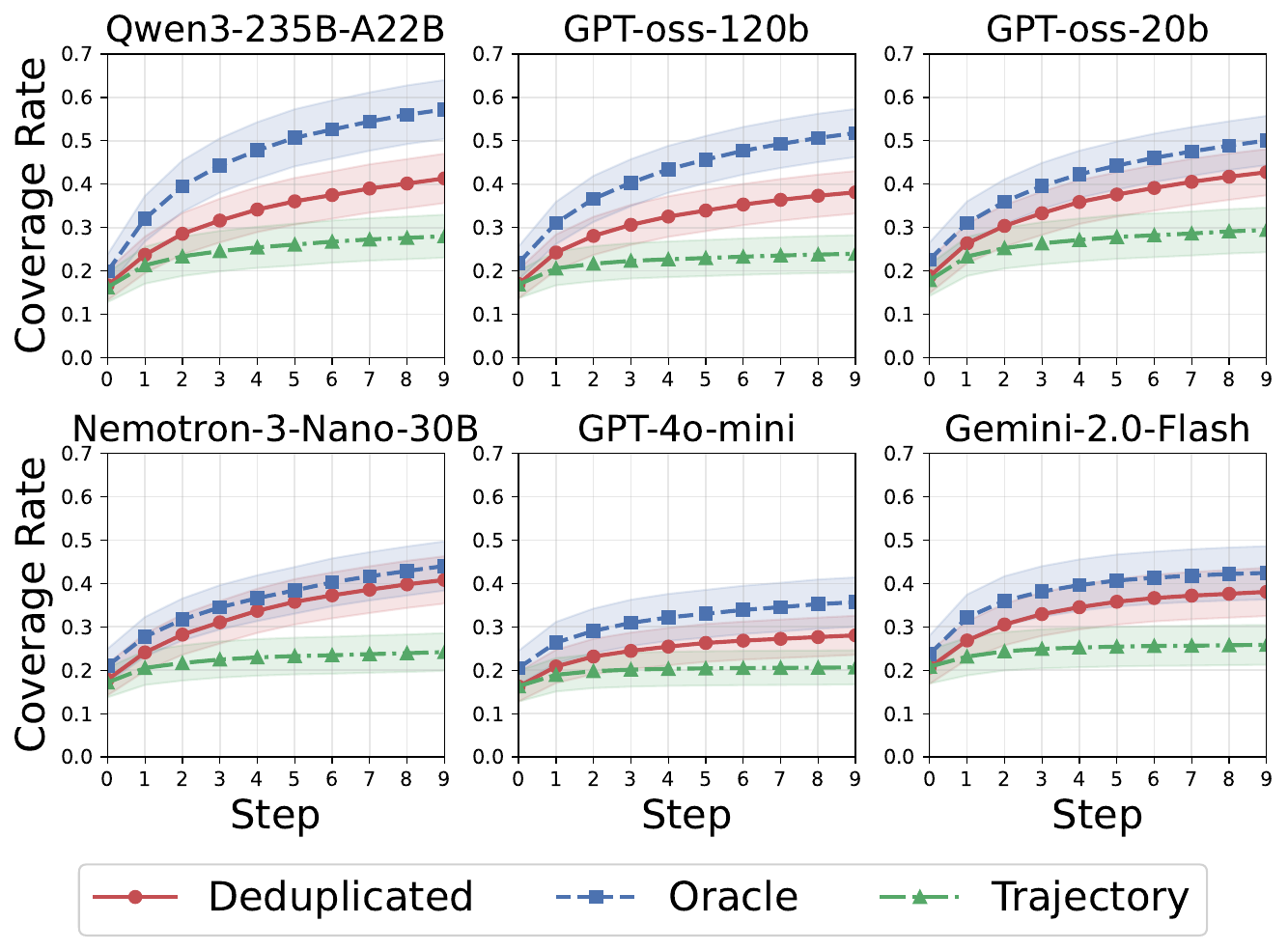}
\caption{Completeness ratio over exploration steps on Wikipedia for each model, broken down by belief representation. Shaded bands show $\pm 0.3$ std across episodes. Across models, the stable signal is the rank ordering raw trajectory $<$ deduplicated trajectory $<$ oracle belief rather than a fixed gap size.}
\label{fig:recall_steps_all}
\end{figure}

\begin{figure}[ht]
\centering
\includegraphics[width=0.9\linewidth]{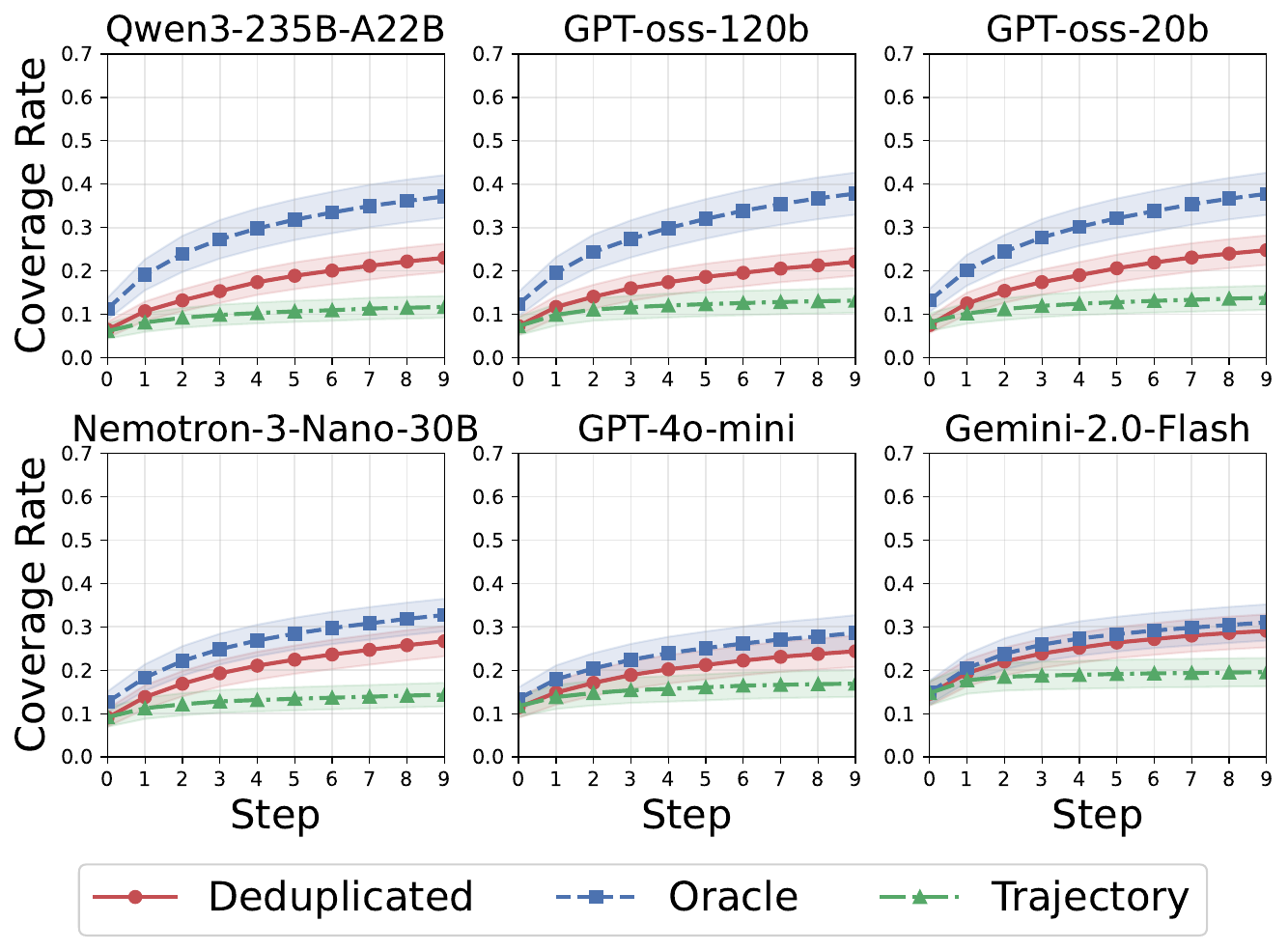}
\caption{Completeness ratio over exploration steps on ML Surveys for each model, broken down by belief representation. Shaded bands show $\pm 0.3$ std across episodes. The same raw trajectory $<$ deduplicated trajectory $<$ oracle belief ordering reappears across models, supporting the generalization of belief-format effects beyond Wikipedia.}
\label{fig:ml_survey_recall_steps_all}
\end{figure}

Figure~\ref{fig:ml_survey_recall_and_cost} provides the appendix-only
ML Surveys counterpart to the main-text Figure~\ref{fig:recall_and_cost}.
Figure~\ref{fig:ml_survey_topic_difficulty} provides the appendix-only
ML Surveys counterpart to the main-text Wikipedia topic-difficulty
heatmap. Figure~\ref{fig:recall_bar_per_model} shows the Wikipedia
final coverage breakdown by model and belief representation, and
Figure~\ref{fig:recall_steps_all} shows the corresponding Wikipedia
step curves underlying aggregated Figure~\ref{fig:recall_and_cost}(a)
in the main paper. Figures~\ref{fig:ml_survey_recall_bar_per_model}
and~\ref{fig:ml_survey_recall_steps_all} show the corresponding ML
Surveys breakdowns.

\subsection{Context Length and Query Diversity}

To quantify whether an agent generates novel queries over time, we
define a per-query diversity score. Let $Q_{<t}$ denote all queries
issued at steps prior to step $t$. For each query $q_i$ at step $t$,
we measure its dissimilarity to the most similar prior query:
\begin{align*}
\text{diversity}(q_i) = 1 - \max_{q \in Q_{<t}} \cos(\text{embed}(q_i),\, \text{embed}(q))
\end{align*}
The step-level diversity is the mean over all queries at that step.
At step~0 no prior queries exist, so we compute pairwise diversity
among queries within the step. A diversity near zero indicates the
agent is repeating queries it has already issued.

\begin{figure}[ht]
\centering
\includegraphics[width=\linewidth]{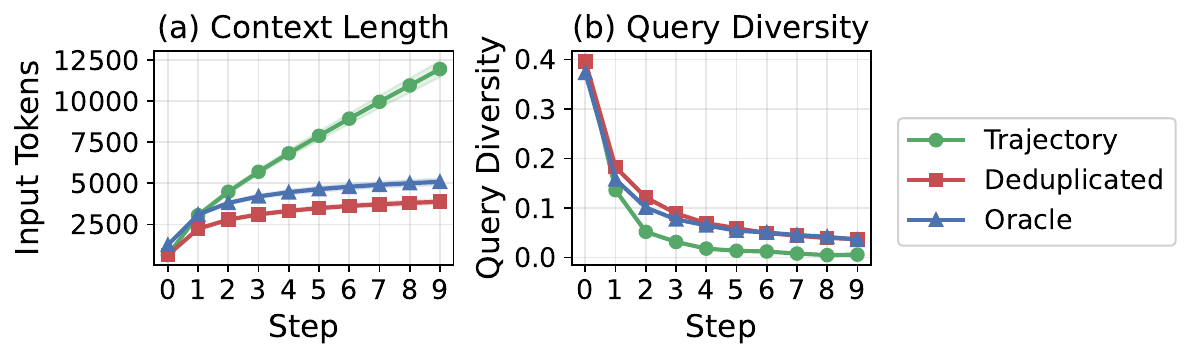}
\caption{Why raw trajectory underperforms on Wikipedia.
(a)~\textit{Context length} over exploration steps, averaged across all
6 models (shaded bands: $\pm 0.3$ std): raw trajectory grows nearly
linearly, while deduplicated trajectory and oracle belief remain more compact.
(b)~\textit{Query diversity} over steps. This more directly reflects
raw trajectory's early coverage saturation: diversity drops sharply over
steps 0--2, then decays at a similar rate but from a lower offset.
deduplicated trajectory and oracle belief have similar diversity despite
oracle belief's higher coverage, indicating that novelty alone does not
explain the oracle belief gain.}
\label{fig:raw_traj_analysis}
\end{figure}

\begin{figure}[ht]
\centering
\includegraphics[width=\linewidth]{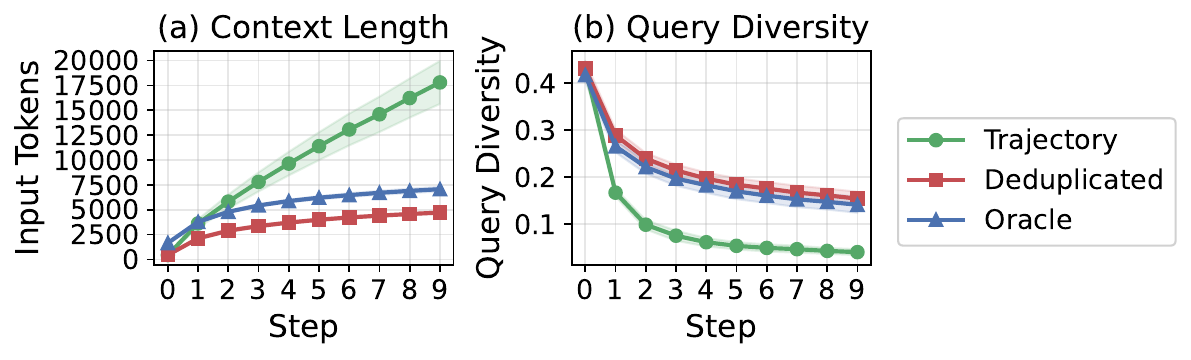}
\caption{Why raw trajectory underperforms on ML Surveys.
(a)~\textit{Context length} over exploration steps, averaged across all
models: raw trajectory again grows nearly linearly, while deduplicated
trajectory and oracle belief remain more compact.
(b)~\textit{Query diversity} over steps. The same pattern appears more
clearly on ML Surveys: raw-trajectory diversity drops sharply over steps 0--2,
then decays at a similar rate but with a larger persistent gap.
deduplicated trajectory and oracle belief again have similar diversity despite
oracle belief's higher coverage, indicating that novelty alone does not
explain the oracle belief gain.}
\label{fig:ml_survey_raw_traj_analysis}
\end{figure}

\subsection{Content Filter Analysis}
\label{app:content_filter}

On Wikipedia, Qwen3-235B-A22B's relatively low coverage on the Chinese History cluster (Figure~\ref{fig:topic_difficulty}) is largely attributable to Alibaba Cloud's content moderation filter. Table~\ref{tab:content_filter} shows that three Chinese political figures (Jiang Zemin, Hu Yaobang, and Li Peng) are blocked at 100\% of steps, yielding zero coverage. Two additional articles (Chen Yun, Zhu Rongji) are partially blocked. In total, 97.9\% of all content filter blocks for Qwen3-235B-A22B occur within the Chinese History cluster, making this an API-level constraint rather than a model capability limitation.

\begin{table}[t]
\caption{Wikipedia articles blocked by Alibaba Cloud's content filter (Qwen-3-235B-A22B). The model returns an ``inappropriate content'' error instead of generating queries. Across all 10,795 agent steps, 338 (3.1\%) were blocked; 97.9\% of blocks occurred in the Chinese History cluster.}
\label{tab:content_filter}
\centering
\small
\begin{tabular}{llrrrl}
\toprule
Article & Topic & Blocked & Total & Block Rate & Avg Coverage Rate \\
\midrule
Jiang Zemin & Chinese History & 90 & 90 & \cellcolor{red!40}100.0\% & 0.000 \\
Hu Yaobang & Chinese History & 90 & 90 & \cellcolor{red!40}100.0\% & 0.000 \\
Li Peng & Chinese History & 90 & 90 & \cellcolor{red!40}100.0\% & 0.000 \\
Chen Yun & Chinese History & 46 & 90 & \cellcolor{red!20}51.1\% & 0.248 \\
Zhu Rongji & Chinese History & 15 & 90 & \cellcolor{red!6}16.7\% & 0.372 \\
Maus & Fiction & 6 & 90 & \cellcolor{red!2}6.7\% & 0.419 \\
Kangaroo & Mammals & 1 & 90 & \cellcolor{red!0}1.1\% & 0.378 \\
\bottomrule
\end{tabular}
\end{table}

\section{Completeness Uncertainty Quantification Analysis}
\label{app:termination_prediction}

\subsection{Conformal Prediction Background and Calibration Algorithm}
\label{app:calibration_algorithm}
\label{app:conformal_background}

Conformal prediction transforms arbitrary uncertainty scores into prediction
sets with finite-sample coverage guarantees. Given a calibration set and a
nonconformity score, conformal prediction computes a threshold such that
prediction sets contain the true answer with specified probability (e.g.,
90\%), without distributional assumptions.

\citet{kumar2023conformal} apply conformal prediction to LLM multiple-choice
questions using softmax scores as nonconformity measures.
\citet{quach2023conformal} extend this to open-ended generation, constructing
prediction sets over semantic clusters. \citet{mohri2024language} provide
tighter bounds through conditional calibration. \citet{angelopoulos2021gentle}
survey the broader framework and its applications. Coverage guarantees hold
regardless of the base predictor's quality; weaker uncertainty estimates yield
larger prediction sets. In our setting, conformal prediction calibrates raw
predicted completeness ratios into intervals that achieve reliable coverage
across models.

Algorithm~\ref{alg:calibration} details the offline calibration
procedure. Given a set of synthetic belief states with known completeness
ratios, it computes nonconformity scores and derives the calibration
adjustment $\hat{q}$ used at inference time.

\begin{algorithm}[H]
\caption{Conformal Calibration (offline, run before Algorithm~\ref{alg:evaluation})}
\label{alg:calibration}
\begin{algorithmic}[1]
\REQUIRE Calibration beliefs $\mathcal{S}_{\text{cal}} = \{(b_s, c_s)\}$ where $b_s$ is a belief state and $c_s = |\textsc{Retrieved}_s| / N$ is the true completeness ratio, estimation method $\mathcal{E}$, coverage level $1-\alpha$
\FOR{each belief $(b_s, c_s) \in \mathcal{S}_{\text{cal}}$}
    \STATE $\hat{c}_s \gets \mathcal{E}(b_s)$ \COMMENT{Estimate completeness ratio}
    \STATE $e_s \gets |c_s - \hat{c}_s|$ \COMMENT{Nonconformity score}
\ENDFOR
\STATE $n \gets |\mathcal{S}_{\text{cal}}|$
\STATE $\hat{q} \gets \lceil(n+1)(1-\alpha)\rceil / n$ quantile of $\{e_s\}_{s=1}^{n}$
\RETURN Calibration adjustment $\hat{q}$ \COMMENT{Prediction interval: $[\hat{c} - \hat{q},\; \hat{c} + \hat{q}]$}
\end{algorithmic}
\end{algorithm}

\subsection{Pre-Calibration Diagnostic Plots}

\begin{figure}[ht]
\centering
\includegraphics[width=0.95\linewidth]{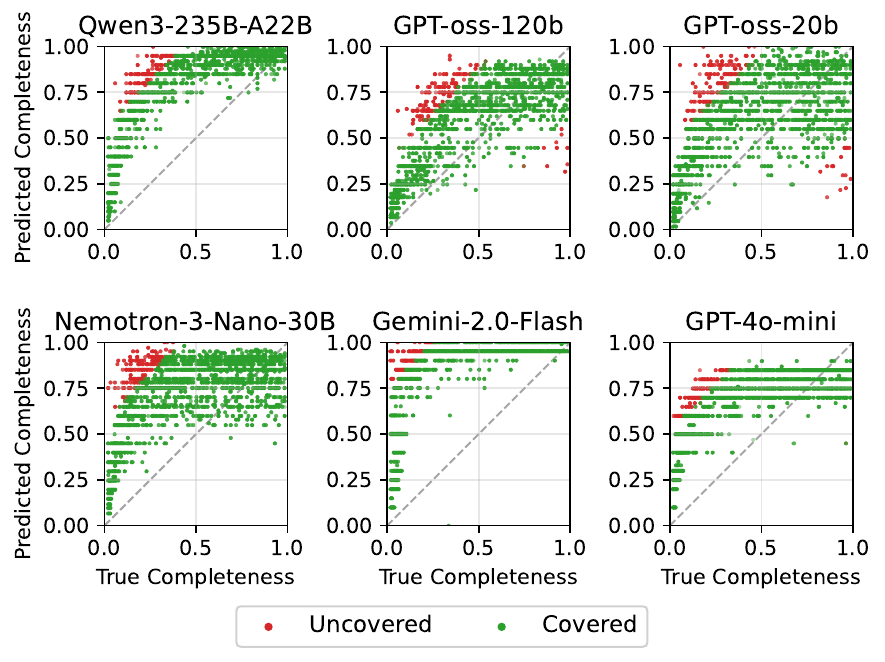}
\caption{Completeness estimation on Wikipedia: true completeness ratio (x-axis) vs.\ predicted completeness ratio (y-axis) per model. Points closer to the diagonal indicate stronger estimators. Across models, predictions tend to overestimate completeness for partially explored articles, especially at intermediate completeness levels. Green points fall within the conformal interval; red points are uncovered.}
\label{fig:scatter_direct_grid}
\end{figure}

\begin{figure}[ht]
\centering
\includegraphics[width=0.95\linewidth]{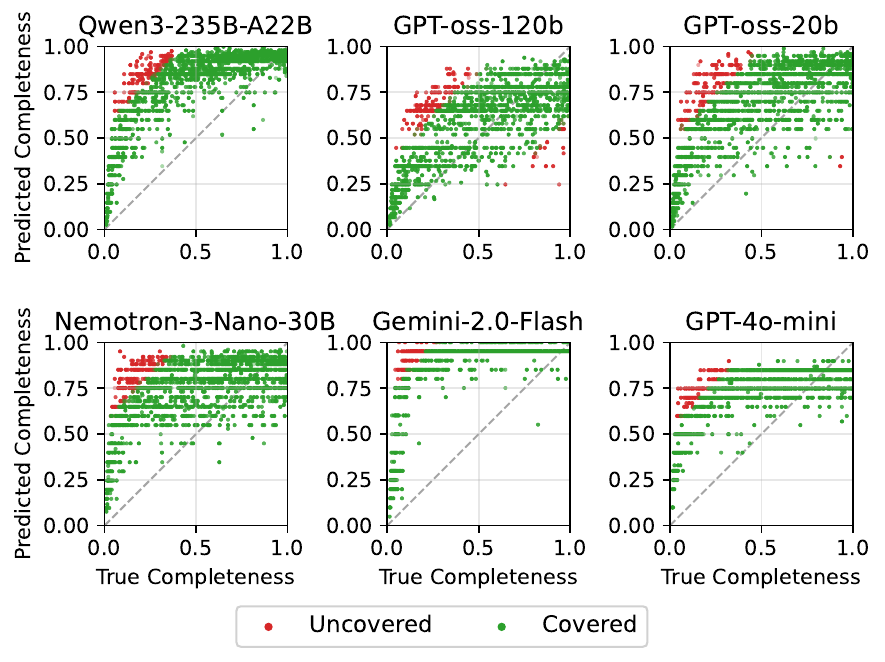}
\caption{Completeness estimation on ML Surveys: true completeness ratio (x-axis) vs.\ predicted completeness ratio (y-axis) per model. Points closer to the diagonal indicate stronger estimators. As on Wikipedia, predictions tend to overestimate completeness for partially explored articles, especially at intermediate completeness levels. Green points fall within the conformal interval; red points are uncovered.}
\label{fig:ml_survey_scatter_direct_grid}
\end{figure}

\begin{figure}[ht]
\centering
\includegraphics[width=0.95\linewidth]{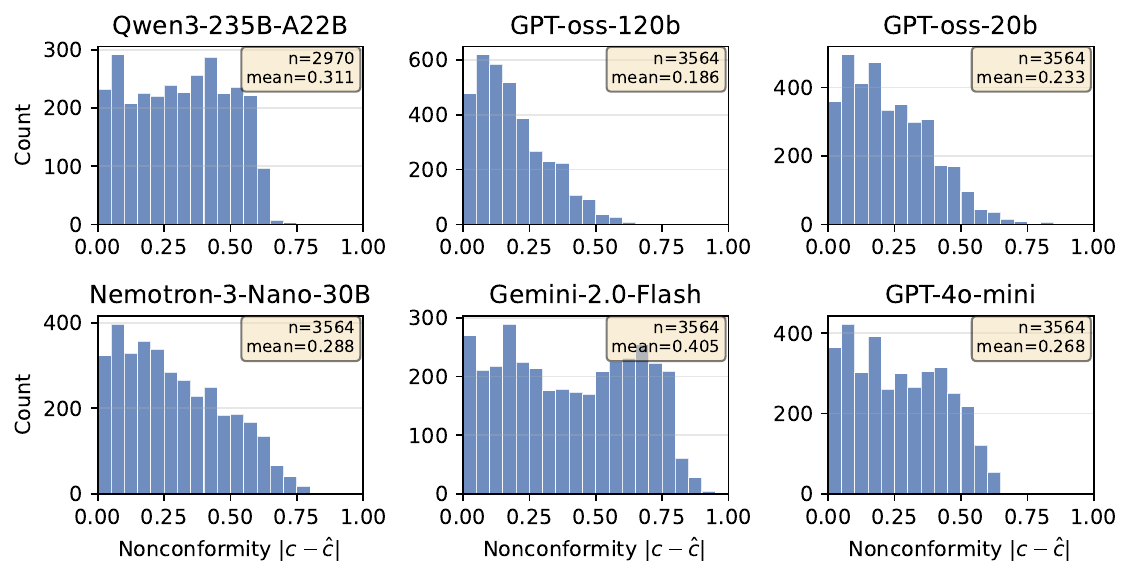}
\caption{Wikipedia nonconformity-score distributions used for conformal calibration, shown separately for each model. Mass concentrated near zero indicates smaller absolute completeness-estimation error, while broader upper tails imply larger calibration correction. The upper quantile of each distribution determines the conformal half-width $\hat{q}$.}
\label{fig:nonconformity_histogram_grid}
\end{figure}

\begin{figure}[ht]
\centering
\includegraphics[width=0.95\linewidth]{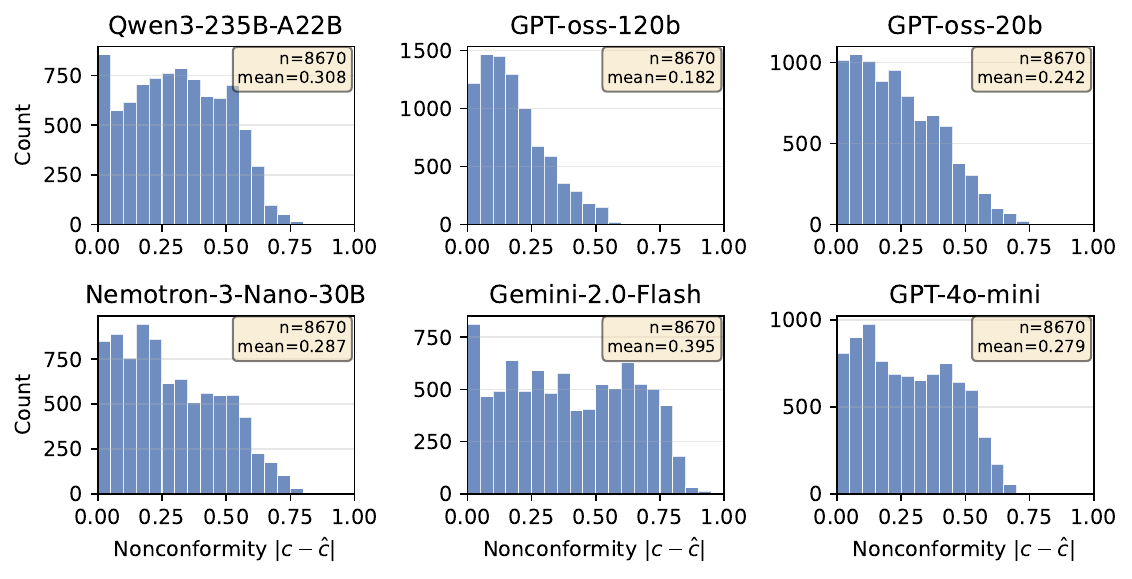}
\caption{ML Surveys nonconformity-score distributions used for conformal calibration. As on Wikipedia, mass concentrated near zero indicates smaller absolute estimation error, while broader upper tails imply larger conformal correction and hence larger $\hat{q}$.}
\label{fig:ml_survey_nonconformity_histogram_grid}
\end{figure}

Figure~\ref{fig:scatter_direct_grid} shows true completeness ratio (x-axis)
vs.\ predicted completeness ratio (y-axis) on Wikipedia for each model before
conformal calibration. Figure~\ref{fig:ml_survey_scatter_direct_grid} shows
the corresponding ML Surveys results. Green points fall within the conformal
interval; red points are uncovered. Figures~\ref{fig:nonconformity_histogram_grid}
and~\ref{fig:ml_survey_nonconformity_histogram_grid} show the corresponding
distributions of nonconformity scores $e = |c - \hat{c}|$ used during
calibration: tighter concentration near zero implies less conformal correction.

\section{Prompt Templates}
\label{app:prompts}

All prompts use a shared XML-tag structure: a \texttt{<role>} block sets the agent's objective, a \texttt{<context>} block provides the article topic and abstract, an optional \texttt{<belief>} (or \texttt{<article>}) block contains the current observation, and an \texttt{<instruction>} block specifies the task. Placeholders in {\color{orange!70!black}\texttt{\{orange monospace\}}} are filled at runtime.

\subsection{Query Generation Prompts}

The system prompt is shared across all belief representations.

\begin{promptbox}[teal!70!black]{System Prompt (all belief types)}
You are an information-seeking agent that generates precise search queries.
Do not include action words like "search", "find", "retrieve", "look up",
or XML tags and structural metadata in queries.
\end{promptbox}

\medskip
The user-turn prompt differs by belief type. At step~0 (no passages collected yet),
all belief types share the same initial prompt:

\begin{promptbox}[teal!70!black]{Query Generation: Initial (step 0, all belief types)}
<role>
\ind Your goal is to retrieve relevant information to complete
\ind an article about "\pvar{topic}".
</role>

<context>
\ind <topic>\pvar{topic}</topic>
\ind <abstract>\pvar{abstract}</abstract>
</context>

<instruction>
\ind Analyze what information is missing or logically incomplete.
\ind Generate up to \pvar{num\_queries} natural language search queries
\ind to retrieve the missing information.
</instruction>
\end{promptbox}

\medskip
From step~1 onward, the belief block is appended. The \textbf{deduplicated trajectory}
representation inserts retrieved passages de-duplicated by content:

\begin{promptbox}[teal!70!black]{Query Generation: deduplicated trajectory followup (steps 1+)}
<role>
\ind Your goal is to retrieve relevant information to complete
\ind an article about "\pvar{topic}".
</role>

<context>
\ind <topic>\pvar{topic}</topic>
\ind <abstract>\pvar{abstract}</abstract>
</context>

<belief>
\ind <passage section="\pvar{section\_name}">
\ind\ind \pvar{passage\_text}
\ind </passage>
\ind ...
</belief>

<instruction>
\ind The found section shows retrieved passages.
\ind
\ind Format:
\ind - <passage section="...">text</passage> shows a retrieved
\ind\ind passage with its section name
\ind
\ind Analyze what information is missing or logically incomplete.
\ind Generate up to \pvar{num\_queries} natural language search queries
\ind to retrieve the missing information.
</instruction>
\end{promptbox}

\medskip
The \textbf{oracle belief} representation replaces the belief block with the full article
skeleton, marking found passages by content and unfound passages with
\texttt{<missing id="pN"/>} gap tags:

\begin{promptbox}[teal!70!black]{Query Generation: oracle belief followup (steps 1+)}
<role>
\ind You have access to the article structure. Your goal is to retrieve
\ind all \pvar{total\_passages} passages to complete an article about
\ind "\pvar{topic}".
</role>

<context>
\ind <topic>\pvar{topic}</topic>
\ind <abstract>\pvar{abstract}</abstract>
</context>

<article>
\ind \pvar{article\_skeleton}
</article>

<instruction>
\ind You have access to the article structure showing all sections
\ind and passage positions.
\ind
\ind Structure format:
\ind - <section name="SectionName"> shows a revealed section
\ind - <section name="???"> shows a hidden section (no passages found)
\ind - <passage id="pN">text</passage> shows a found passage
\ind - <missing id="pN"/> shows a passage you still need to find
\ind
\ind Analyze what information is missing or logically incomplete.
\ind Generate up to \pvar{num\_queries} natural language search queries
\ind to retrieve the missing information.
</instruction>
\end{promptbox}

\medskip
The \textbf{raw trajectory} representation appends the full query--result history
instead of de-duplicated passages:

\begin{promptbox}[teal!70!black]{Query Generation: raw trajectory followup (steps 1+)}
<role>
\ind Your goal is to retrieve relevant information to complete
\ind an article about "\pvar{topic}".
</role>

<context>
\ind <topic>\pvar{topic}</topic>
\ind <abstract>\pvar{abstract}</abstract>
</context>

<belief>
\ind <query\_result>
\ind\ind <query>\pvar{query\_text}</query>
\ind\ind <passages>
\ind\ind\ind <passage section="\pvar{section}">
\ind\ind\ind\ind \pvar{passage\_text}
\ind\ind\ind </passage>
\ind\ind\ind ...
\ind\ind </passages>
\ind </query\_result>
\ind ...
</belief>

<instruction>
\ind The found section shows your search history as query-result pairs.
\ind
\ind Format:
\ind - <query\_result> contains a query and its results
\ind - <query>...</query> shows a search query you made
\ind - <passages> contains the retrieved passages
\ind - <passage section="...">text</passage> shows a retrieved passage
\ind - <no\_results/> indicates a query found no results
\ind
\ind Analyze what information is missing or logically incomplete.
\ind Generate up to \pvar{num\_queries} natural language search queries
\ind to retrieve the missing information.
</instruction>
\end{promptbox}

\subsection{Completeness Uncertainty Quantification Prompts}

For completeness uncertainty quantification, the belief block \pvar{formatted\_passages} is always
formatted using the deduplicated trajectory representation (de-duplicated \texttt{<passage>}
tags in document order).

\paragraph{Completeness-Ratio Estimation.}
The model is prompted to estimate the completeness fraction of the currently
collected passages as a scalar between 0 and 1.

\begin{promptbox}[orange!70!black]{Completeness-Ratio Estimation: Point Estimate}
<role>
\ind You are evaluating information retrieved for a target document
\ind about "\pvar{topic}".
</role>

<context>
\ind <topic>\pvar{topic}</topic>
\ind <abstract>\pvar{abstract}</abstract>
</context>

<belief>
\ind \pvar{formatted\_passages}
</belief>

<instruction>
\ind You have collected \pvar{num\_collected} passages so far from the
\ind target document.
\ind
\ind Estimate the completeness fraction of the current collection between
\ind 0 and 1, where:
\ind - 0 = you have found essentially none of the document's passages
\ind - 0.5 = you have found about half of the document's passages
\ind - 1 = you have found all or almost all of the document's passages
\ind
\ind Return a JSON object with:
\ind - predicted\_completeness: your best estimate of the completeness
\ind   fraction as a number from 0 to 1
\ind
\ind predicted\_completeness should be a direct estimate of the
\ind completeness fraction, not a vague score.
\ind Use the full range from 0 to 1 when appropriate.
\ind Base your estimate on the passages collected so far, the abstract,
\ind and any obvious missing coverage.
</instruction>
\end{promptbox}

\section{Hyperparameters}
\label{app:hyperparameters}

Each episode allows $K = 10$ queries per step over $M = 10$ steps, for a total query budget of 100 queries per episode. The retrieval similarity threshold is $\theta = 0.65$ and the conformal coverage target is $1 - \alpha = 90\%$. All models are queried with temperature 0.7 and a maximum of 8{,}192 output tokens. We use structured JSON output to enforce consistent query formatting; Qwen3-235B-A22B is an exception, as its API provider does not support structured output, so queries are parsed from free-form text. For GPT-oss models we additionally set \texttt{reasoning\_effort=medium} to control chain-of-thought length and API cost.

\subsection*{Evaluation Protocol}
\label{app:evaluation_protocol}

Algorithm~\ref{alg:evaluation} formalizes the full evaluation loop. For
each test document, the agent iterates through query generation,
retrieval, and belief updates until the calibrated stopping criterion is
met or the step budget is exhausted.

\begin{algorithm}[H]
\caption{SeekerGym Evaluation Protocol}
\label{alg:evaluation}
\begin{algorithmic}[1]
\REQUIRE Agent $\pi$, test documents $\mathcal{D}_{\text{test}}$, embedding model $\text{embed}(\cdot)$, similarity threshold $\theta$, queries per step $K$, max steps $M$, belief representation $\mathcal{B} \in \{\text{raw trajectory}, \text{deduplicated trajectory}, \text{oracle belief}\}$, estimation method $\mathcal{E}$, stopping threshold $\delta$, calibration adjustment $\hat{q}$ (from Algorithm~\ref{alg:calibration})
\STATE $\textsc{CompletenessRatios} \gets []$
\FOR{each document $d \in \mathcal{D}_{\text{test}}$}
    \STATE $\mathcal{X}^{\text{goal}} \gets \text{paragraphs}(d)$, \; $b_0 \gets \text{abstract}(d)$, \; $\textsc{Retrieved} \gets \emptyset$
    \FOR{$t = 1$ to $M$}
        \STATE $\hat{c} \gets \mathcal{E}(b_{t-1})$ \COMMENT{Estimate completeness ratio}
        \IF{$\hat{c} - \hat{q} \geq \delta$}
            \STATE \textbf{break} \COMMENT{Calibrated lower bound exceeds threshold}
        \ENDIF
        \STATE $\{a_1, \ldots, a_K\} \gets \pi(b_{t-1})$ \COMMENT{Generate $K$ queries}
        \FOR{$i = 1$ to $K$}
            \STATE $o_i \gets \{x \in \mathcal{X}^{\text{goal}} : \cos(\text{embed}(a_i), \text{embed}(x)) > \theta\}$
            \STATE $\textsc{Retrieved} \gets \textsc{Retrieved} \cup\; o_i$
        \ENDFOR
        \STATE $b_t \gets \mathcal{B}(b_{t-1}, \{(a_i, o_i)\}_{i=1}^K)$ \COMMENT{Update belief}
    \ENDFOR
    \STATE $c_d \gets |\textsc{Retrieved}| \;/\; |\mathcal{X}^{\text{goal}}|$
    \STATE Append $c_d$ to $\textsc{CompletenessRatios}$
\ENDFOR
\RETURN $\text{mean}(\textsc{CompletenessRatios})$
\end{algorithmic}
\end{algorithm}

\section{Limitations and Scope}
\label{app:limitations}

\subsection{Domain Coverage}

SeekerGym's benchmark framework generalizes to any passage-structured
corpus; the current instantiation on Wikipedia and ML Surveys
demonstrates this across general-domain and specialized scientific
content, but coverage of additional domains (e.g., legal filings,
clinical records) remains to be validated.

\subsection{Retrieval Mechanism}

SeekerGym currently uses embedding-based vector search as its
retrieval mechanism. We explored BM25 as an alternative but found
that its word-level decomposition introduced excessive noise,
producing inconsistent retrieval results and making it unsuitable
for evaluating natural-language query generation. Extending to
other retrieval paradigms (e.g., hybrid search, structured queries)
remains an open direction.

\subsection{Methodology}

Our completeness uncertainty quantification evaluation uses synthetic belief states
with uniformly sampled passage subsets, which may differ in
distribution from real agent trajectories where retrieval order is
correlated. While the conformal coverage guarantee holds by
construction on the synthetic distribution, calibration transfer to
deployment trajectories is an open question. Additionally, while we
propose a calibrated stopping criterion
($\hat{c} - \hat{q} \geq \delta$), our experiments evaluate
completeness uncertainty quantification offline on fixed belief states; validating
adaptive stopping during live exploration, where estimation errors
may compound across steps, remains an important direction.

\end{document}